\documentclass[review,3p,times,12pt]{elsarticle} 

\usepackage{hyperref}

\usepackage{graphicx}

\DeclareGraphicsExtensions{.eps}

\usepackage{amssymb}
\usepackage[cmex10]{amsmath}
\usepackage{dsfont}

\usepackage{url}
\usepackage{epstopdf}
\usepackage{pdfpages}
\usepackage{colortbl}
\usepackage[ruled,vlined]{algorithm2e}
\usepackage{subfigure}
\usepackage{enumitem}

\makeatletter
\newcommand{\nosemic}{\renewcommand{\@endalgocfline}{\relax}}
\newcommand{\dosemic}{\renewcommand{\@endalgocfline}{\algocf@endline}}
\makeatother

\journal{Information Sciences}

\begin{document}

\begin{frontmatter}

\title{A Greedy Search Tree Heuristic for Symbolic Regression}

\author[olivetti]{Fabr\'icio~Olivetti~de~Fran\c{c}a}
\address[olivetti]{Center of Mathematics, Computing and Cognition (CMCC), Universidade Federal do ABC (UFABC) -- Santo Andr\'{e}, SP, Brazil. E-mail: folivetti@ufabc.edu.br}

\begin{abstract}
Symbolic Regression tries to find a mathematical expression that describes the relationship of a set of explanatory variables to a measured variable. The main objective is to find a model that minimizes the error and, optionally, that also minimizes the expression size. A smaller expression can be seen as an interpretable model considered a reliable decision model. This is often performed with Genetic Programming which represents their solution as expression trees. The shortcoming of this algorithm lies on this representation that defines a rugged search space and contains expressions of any size and difficulty. These pose as a challenge to find the optimal solution under computational constraints. This paper introduces a new data structure, called Interaction-Transformation (IT), that constrains the search space in order to exclude a region of larger and more complicated expressions. In order to test this data structure, it was also introduced an heuristic called SymTree. The obtained results show evidence that SymTree are capable of obtaining the optimal solution whenever the target function is within the search space of the IT data structure and competitive results when it is not. Overall, the algorithm found a good compromise between accuracy and simplicity for all the generated models.
\end{abstract}

\begin{keyword}
Symbolic Regression \sep Regression Analysis \sep Greedy Heuristic
\end{keyword}

\end{frontmatter}

\section{Introduction}
\label{sec:introduction}

Many decision making process can be automated by learning a computational model through a set of observed data. For example, credit risk can be estimated by using explanatory variables related to the consumer behavior~\cite{crouhy2000comparative}. A recommender system can estimate the likelihood of a given person to consume an item given their past transactions~\cite{adomavicius2005toward}. 

There are many techniques devised to generate such models, from the simple Linear Regression~\cite{seber2012linear} to more advanced universal approximators like Neural Networks~\cite{haykin2004comprehensive}. The former has the advantage of being simple and easily interpretable, but the relationship must be close to linear for the approximation to be acceptable. The latter can numerically approximate any function given the constraint that the final form of the function is pre-determined, usually as a weighted sum of a nonlinear function applied to a linear combination of the original variables. This constraint makes the regression model hard to understand, since the implications of changing the value of an input variable is not easily traced to the target variable. Because of that, these models are often called \textit{Black Box Models}.

The concerns with using black box models for decision making process are the inability to predict what these models will do in critical scenarios and whether their response are biased by the data used to adjust the parameters of the model. 

For example, there is recent concern on how driverless cars will deal with variants of the famous Trolley problem~\cite{bonnefon2015autonomous, thomson1976killing}. Driverless cars use classification and regression model to decide the next action to perform, such as speed up, slow down, break, turn left or right by some degrees, etc. If the model used by these cars are difficult to understand, the manufacturer cannot be sure the actions the car will do in extreme situations. Faced with the decision of killing a pedestrian or killing the driver, what choice will it make? Even though these situations may be rare, it is important to understand whether the model comprehends all possible alternatives to prevent life losses.

Another recent example concerns the regression models used to choose which online ad to show to a given user. It was found in~\cite{datta2015automated} that the model presented a bias towards the gender of the user. Whenever the user was identified as a male, the model chose ads of higher paying jobs than when the user was a female person. In this case, the bias was introduced by the data used as a reference to adjust the parameters of the model. Historically, the income distribution of men is skewed towards higher salaries than women~\cite{suter1973income}.

An interpretable model could provide a better insight to such concerns since everything will be explicitly described in the mathematical expression of the model. In the example of the driverless car, an inspection on the use of variables corresponding to location of bystanders around the car could reveal what would be the probable actions taken by the model. Simlarly, the inspection the mathematical expression to choose the online ads, could reveal a negative correlation for the combination of salary and the female gender.

As such, an interpretable model should have both high accuracy regarding the target variable and, at the same time, be as simple as possible to allow the interpretation of the decision making process.

Currently this type of model is being studied through \emph{Symbolic Regression}~\cite{billard2002symbolic}, a field of study that aims to find a symbolic expression that fits an examplary data set accurately. Often, it is also included as a secondary objective that such expression is as simple as possible. This is often solved by means of Genetic Programming~\cite{koza1992genetic}, a metaheuristic from the Evolutionary Algorithms~\cite{back1996evolutionary} field that evolves an expression tree by minimizing the model error and maximizing the simplicity of such tree. Currently, the main challenges in such approach is that the search space induced by the tree representation not always allow a smooth transition between the current solution towards an incremental improvement and, since the search space is unrestricted, it allows the representation of black box models as well.

\subsection{Objectives and Hypothesis}

This main objective of this paper is to introduce a new data structure, named \emph{Interaction-Transformation} (IT), for representing mathematical expressions that constrains the search space by removing the region comprising uninterpretable expressions. Additionaly, a greedy divisive search heuristic called \emph{SymTree} is proposed to verify the suitability of such data structure to generate smaller Symbolic Regression models.

The data structure simply describes a mathematical expression as the summation of polynomial functions and transformation functions applied to the original set of variables. This data structure restrict the search space of mathematical expressions and, as such, is not capable of representing every possible expression.

As such, there are two hypothesis being tested in this paper:

\begin{enumerate}[label=\textbf{H\arabic*.}]

\item The IT data structure constrain the search space such as it is only possible to generate smaller expressions.

\item Even though the search space is restricted, this data structure is capable of finding function approximations with competitive accuracy when compared to black box models.

\end{enumerate}

In order to test these hypothesis the \emph{SymTree} algorithm will be applied to standard benchmark functions commonly used on the literature. These functions are low dimensional functions but that still proves to be a challenge to many Symbolic Regression algorithms. The functions will be evaluated by means of Mean Squared Error and number of nodes in the tree representation of the generated expression. Finally, these results will be compared to three standard regression approaches (linear and nonlinear), three recent variations of Genetic Programming applied to this problem and two other Symbolic Regression algorithms from the literature.

The experimental results will show that the proposed algorithm coupled with this data structure is indeed capable of finding the original form of the target functions whenever the particular function is representable by the structure. Also, when the function is not representable by the IT data structure, the algorithm can still manage to find an approximation that compromises between simplicity and accuracy. Regarding numerical results, the algorithm performed better than the tested Symbolic Regression algorithms in most benchmarks and it was competitive when compared against an advanced black box model extensively used on the literature.

The remainder of this paper is organized as follows, Section~\ref{sec:symreg} gives a brief explanation of Symbolic Regression and classical solution through Genetic Programming. In Section~\ref{sec:review} some recent work on this application is reported along with their contribution. Section~\ref{sec:proposal} describes the proposed algorithm in detail, highlighting its advantages and limitations. Section~\ref{sec:experiments} explains the experiments performed to assess the performance of the proposed algorithm and compare its results with the algorithms described in Section~\ref{sec:review}. Finally, Section~\ref{sec:conclusion} summarizes the contributions of this work and discuss some of the possibilities for future research.

\section{Symbolic Regression}
\label{sec:symreg}


Consider the problem where we have collected $n$ data points $X = \{\mathbf{x_1},...,\mathbf{x_n}\}$, called explanatory variables, and a set of $n$ corresponding target variables $Y = \{y_1,...,y_n\}$. Each data point is described as a vector with $d$ measurable variables $\mathbf{x_i} \in \mathbb{R}^d$. The goal is to find a function $\hat{f}(\mathbf{x}): X \rightarrow Y$, also called a model, that approximates the relationship of a given $\mathbf{x_i}$ with its corresponding $y_i$.

Sometimes, this can be accomplished by a linear regression where the model is described by a linear function assuming that the relationship between the explanatory and target variables are linear. When such assumption does not hold, we can use non-linear regression techniques, such as Artificial Neural Network, which have theoretical guarantees to the capability of approximating any given function. The problem with the latter is that every function has the form:

\begin{equation}
\hat{f}(\mathbf{x}) = \sum_{i=1}^{K}{v_k \cdot g(\mathbf{w_i^T} \cdot \mathbf{x} + b_i)},
\end{equation}

\noindent where $\mathbf{v}$ is a vector representing the regression coefficients for the nonlinear function $g(.)$, $W$ is a $K \times d$ matrix containing the linear regression coefficients for every explanatory variable respective to each nonlinear function, $\mathbf{b}$ is a vector with the biases for each of the $K$ linear regressions and $g(.)$ is any non-constant, bounded, monotonically-increasing, continuous nonlinear function. It is proven~\cite{gybenko1989approximation, hornik1991approximation} that given the correct values of $K$, $\mathbf{v}$, $W$ and $\mathbf{b}$, any function $f(.)$ can be approximated with a small error $\epsilon$.

If the goal of a given problem is just to obtain the numerical approximation of such function, this nonlinear model may suffice. But, if there is a need for understanding, inspecting or even complementing the obtained model this approximation becomes impractical since this function form is fixed and only the coefficients are optimized. 

In these situations we need to find not only a function $\hat{f}$ that approximates the observed relationship by minimizing the error but also one that maximizes the simplicity or interpretability. This is often achieved by means of Symbolic Regression. 

In Symbolic Regression both the function form and the coefficients are searched with the objective of minimizing the approximation error and, in some cases, also maximizing the simplicity of the expression. The meaning of \textit{simplest} in the context of Symbolic Regression refers to the ease of interpretation. For example, consider the following functions:

\begin{align*}
f(x) &= \frac{x^3}{6} + \frac{x^5}{120} + \frac{x^7}{5040} \\
f(x) &= \frac{16x(\pi - x)}{5\pi^2 - 4x(\pi - x)} \\
f(x) &= \sin{(x)}
\end{align*}

Assuming that the third function is the target, the first two functions return a reasonable approximation within a limited range of the $x$ values. If we have the goal to interpret the function behavior, instead of just obtaining the numerical approximation, the target function is the simplest form and the simplest to understand. The simplicity is often measured by means of the size of the expression tree or the number of nonlinear functions used in the expression.

\begin{figure}
\centering
\includegraphics[height=0.3\textheight]{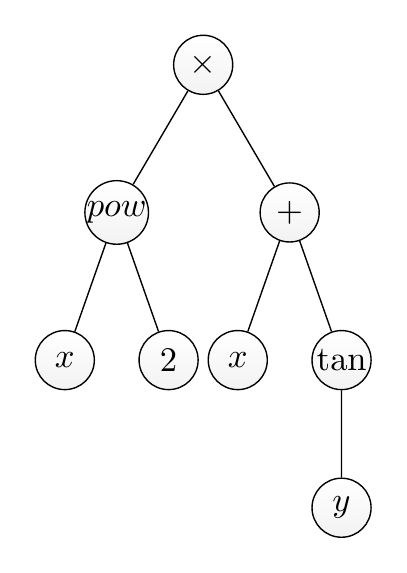}
\caption{Expression tree for the expression $x^2 \cdot (x + \tan{(y)})$.}
\label{fig:exptree}
\end{figure}

The expression tree is a tree data structure representing a given expression. Each node of the tree may represent an operator, a function, a variable or a constant. The nodes representing an operator or a function must have a set of child nodes for each of the required parameters. The variables and constants should all be leaf nodes. For example, the expression $x^2 \cdot (x + \tan{y})$ can be represented by the tree depicted in Fig.~\ref{fig:exptree}. In this example the length of the expression could be measured by the number of nodes ($8$) or the height of the three ($3$). Additionally, the simplicity can be measured by penalizing the \emph{bloatness}~\footnote{In GP the term \emph{bloat} is used to refer to large and complicated expressions} of the expression for the number of nonlinear functions used, the size of the composition of functions and the total number of variables used.

This expression tree is often used as a representation to an evolutionary algorithm called Genetic Programming in order to find such optimal expression for a regression problem.

\subsection{Genetic Programming}
\label{sec:gp}

The Genetic Programming (GP) algorithm, in the context of Symbolic Regression, tries to find the optimal expression that minimizes the approximation error. As a secondary objective it is often sought to also maximizes the simplicity of the expression, through regularization.

As usual for any evolutionary algorithm, GP starts with a population of randomly generated solutions, usually by means of the Ramped half-and-half procedure to improve the variability of the structure~\cite{koza1992genetic}, and afterwards iterates through the procedures of reproduction, mutation and selection. The solutions are represented by means of expression trees.

The reproduction procedure tries to combine the good parts of two or more solutions creating a new and improved solution, this procedure works well whenever a part of the solution representation is translated to a subset of the original problem. 


The mutation, or perturbation, operator is responsible for introducing small changes to a given solution in order to prevent the search process of getting stuck on a local optima. This operator works well whenever a small change to a solution does not change the fitness by a large amount. For example, with numerical optimization the mutation is usually the addition of a random Gaussian vector such that $|f(x) - f(x +\sigma)| < \epsilon$.

\begin{figure}[ht!]
\centering
\subfigure[]{
\includegraphics[height=0.15\textheight]{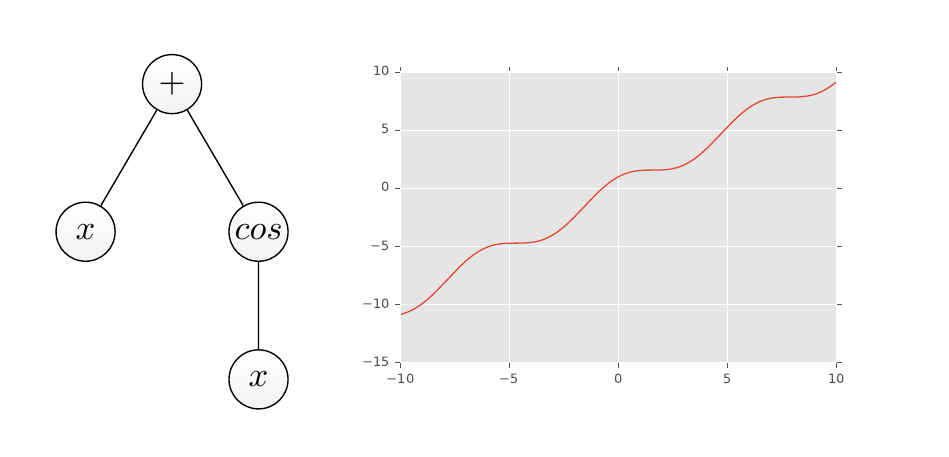}
\label{fig:gpA}
} \\
\subfigure[]{
\includegraphics[height=0.15\textheight]{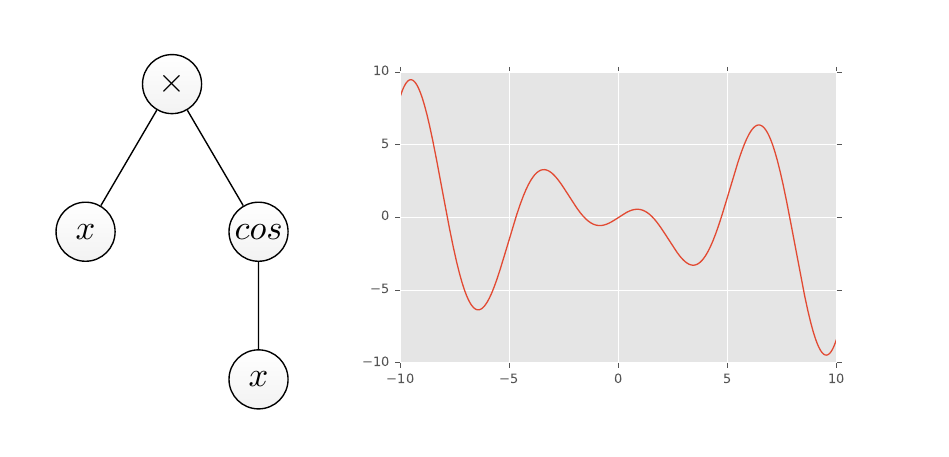}
\label{fig:gpB}
}
\subfigure[]{
\includegraphics[height=0.15\textheight]{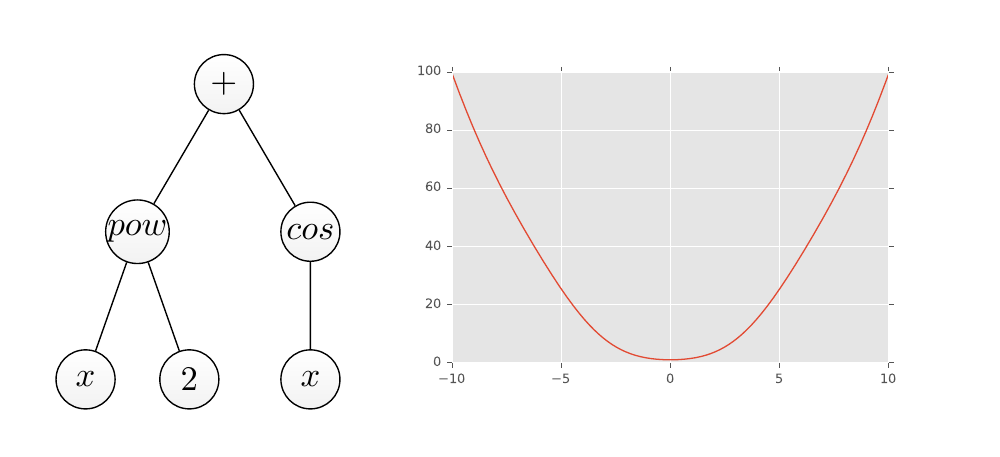}
\label{fig:gpC}
}
\caption{Example of the application of mutation (b) and crossover (c) operators in the expression $x + \cos{(x)}$ (a).}
\label{fig:gpex}

\end{figure}

But, the expression tree representation does not guarantee any of these properties for the common evolutionary operators. For example, given the expression tree in Fig.~\ref{fig:gpA}, a possible mutation may change the addition operator to the multiplication operator, resulting in the expression in Fig.~\ref{fig:gpB}. In Fig.~\ref{fig:gpC} we can see a possible result of the application of the crossover operator, in this situation an entire sub-tree is replaced by a new one. As we can see from the respective plots of each expression they approximate completely different relations.

Many researchers have created specialized operators to alleviate the problem with the standard operators. In the next section, we will summarize some of the most recent proposals in the literature.

\section{Literature Review}
\label{sec:review}

Recently, many extensions to the cannonical GP or even new algorithms were proposed in order to cope with the shortcomings pointed out in the previous section. This section will highlight some of the recent publications that reported to have achieved improvements over previous works.


In~\cite{trujillo2016neat} the authors propose the \textit{neat}-GP, a GP based algorithm with mechanisms borrowed from Neuroevolution of augmenting topologies~\cite{stanley2002evolving} (NEAT) algorithm and the Flat Operator Equalization bloat control method for GP. From the former it borrows the speciation mechanism through fitness sharing~\cite{sareni1998fitness, de2010diversity}, forcing the crossover operator to be applied only on similar solutions and, from the latter, it maximizes the simplicity of generated expression by encouraging an uniform distribution on the tree size among solutions of the population. The authors tested \textit{neat}-GP against the classic GP algorithm and the GP with the Flat Operator Equalization. Additionally, the authors also tested different variations of \textit{neat}-GP by replacing operators, selection and fitness sharing mechanisms. For every tested function, at least one variation of the proposed algorithm achieved the best solution.

Instead of modifying the search operators, in~\cite{kattan2016gp} the authors proposed the use of a surrogate model to reduce the fitness calculation cost, thus allowing a higher number of iterations within a time frame. In their proposal, called Semantic Surrogate Genetic Programming (SSGP), the fitness approximation is composed of a linear equation controlling the number of training samples used to calculate a partial fitness from the expression tree and the application of the k-NN algorithm to infer the remaining training samples. When comparing the results of SSGP against four other Genetic Programming variants the authors showed that their algorithm could maintain a comparable solution, and in some situations even better, than the contenders.

In~\cite{kattan2015surrogate} the authors also explored the use of a surrogate model to alleviate the cost of fitness evaluation. The proposed algorithm, Surrogate Genetic Programming (sGP), trains two Radial Basis Functions Neural Networks~\cite{park1993approximation} (RBFN) in order to first map the semantic space (described by the expression tree) into the output space (composed of the target vector) and then another network maps the output space into the fitness space. The sGP uses the surrogate model with $40\%$ of the population while the remaining solutions are fully evaluated. Comparing the sGP against three other variants, the authors showed that their approach obtained the best results for every tested function using the same number of function evaluations.

The algorithm named Evolutionary Feature Synthesis, introduce in~\cite{arnaldo2015building}, tries to fit a model in the form:

\begin{equation}
\hat{f}(x) = \sum_{i=1}^{M}{w_i \cdot h_i(x)},
\label{eq:efs}
\end{equation}

\noindent that minimizes the squared error. Differently from Genetic Programming, the algorithm tries to evolve a population of $M$ terms instead of a population of expressions, so at every step a new term can be created, from a combination of terms of the current population, and old ones are discarded with probability inversely proportional to $|w_i|$. The function $h_i(x)$ can be any composition of functions with one and two variables. The authors showed that this fixed model form was capable of finding more acurate models when compared to traditional Genetic Programming.


Finally, in~\cite{icke2013improving} the authors expanded the idea of the Fast Function Extraction algorithm~\cite{mcconaghy2011ffx} (FFX) by creating the hybrid FFX/GP algorithm. The FFX algorithm enumerates the binary interactions between the original variables of the problem and some polynomial variations of such variables. After this enumeration, different linear models are created by using the ElasticNet linear regression~\cite{zou2005regularization} which acts as a feature selection mechanism. The ElasticNet is applied with different regularization parameters rendering several different solutions. This set of solutions is then used by FFX/GP to generate a new data set by extending the variable space with every unique interaction found in the set of solutions. The GP algorithm is then applied in this new data set. This algorithm was tested by a huge set of randomly generated polynomial functions and the results showed that FFX/GP improved the capabilities of FFX when dealing with higher order polynomials.


\section{Constrained Representation for Symbolic Regression}
\label{sec:proposal}

The overall idea introduced in this section is that if a given representation for Symbolic Regression does not comprehend \emph{bloated} mathematical expressions, it will allow the algorithms to focus the search only on the subset of expressions that can be interpretable.

For this purpose, a new Data Structure used to represent the search space of mathematical expressions will be introduced followed by an heuristic algorithm that makes use of such representation to find approximations to nonlinear functions.

\subsection{Interaction-Transformation Data Structure}
\label{sec:representation}

Consider the regression problem described in Sec.~\ref{sec:symreg} consisting of $n$ data points described by a $d$-dimensional vector of variables $X$ and corresponding target variables $Y$. The goal is to find the \textit{simplest} function form $\hat{f}: \mathbb{R}^{n} \rightarrow \mathbb{R}$ that minimizes the approximation to the target variables. 

Let us describe the approximation to the target function by means of a linear regression of functions in the form:

\begin{equation}
\hat{f}(x) = \sum_{i}{w_i \cdot g_i(x)},
\label{eq:appfunction}
\end{equation}

\noindent where $w_i$ is the $i$-th coefficient of a linear regression and $g_i(.)$ is the $i$-th function (non-linear or linear). The function $g(.)$ is a composition function $g(.) = t(.) \circ p(.)$, with $t: \mathbb{R} \rightarrow \mathbb{R}$ a one-dimensional transformation function and $p: \mathbb{R}^{d} \rightarrow \mathbb{R}$ a $d$-dimensional interaction function. The interaction function has the form:

\begin{equation}
p(x) = \prod_{i=1}^{d}{x_i^{k_i}},
\end{equation}

\noindent where $k_i \in \mathbb{Z}$ is the exponent of the $i$-th variable.

This approximation function (Eq.~\ref{eq:appfunction}) is generic enough to correctly describe many different functions. Following the previous example, the function $\sin{x}$ can be described as:

\begin{equation*}
\hat{f}(x) = 1 \cdot \sin(x).
\end{equation*}

In this example, there is only one term for the summatory of Eq.~\ref{eq:appfunction}, the single coefficient $w_1 = 1$ and the composition function $g(x)$ is simply the composition of $t(z) = \sin{(z)}$ and $p(x) = x$. As a more advanced example, suppose the target function is $f(x) = 3.5\sin{(x_1^2 \cdot x_2)} + 5\log{(x_2^3/x_1)}$. This function could be described as:

\begin{equation*}
\hat{f}(x) = 3.5 \cdot g_1(x) + 5 \cdot g_2(x),
\end{equation*}

\noindent and

\begin{align*}
t_1(z) &= \sin{(z)} \\
p_1(\mathbf{x}) &= x_1^2 \cdot x_2 \\
t_2(z) &= \log{(z)} \\
p_2(\mathbf{x}) &= x_1^{-1} \cdot x_2^3.
\end{align*}

Notice that Eq.~\ref{eq:appfunction} does not comprehend the description of a more complicated set of mathematical expressions such as  $f(x) = \sin{(\log{(\cos{(x)})})}$, which prevents the search algorithm to find \emph{bloated} expressions. On the other hand this representation also does not comprehend expressions such as $f(x) = \sin{(x_1^2 + x_2)/x_3}$ or $f(x) = \sin{(x_1^2 + 2 \cdot x_2)}$ which can be deemed as simple.

As a comparison with the EFS algorithm, introduced in Sec.~\ref{sec:review}, notice that in the approximation function described in~\ref{eq:efs}, $h_i(x)$ can be any composition of functions without constant values. From the previou paragraph, it means it can represent the first and second functions, but not the third one.

A computational representation of the function described in Eq.~\ref{eq:appfunction} can be derived by a set $T = \{t_1, ..., t_n\}$ of terms. Each term represents a tuple $(P, t)$, with $P = [p_1,...,p_d]$ representing an interaction of the original variables with each variable $i$ having an exponent $p_i$ and $t$ a transformation function. 

As previously indicated, the transformation function may be any mathematical function $t: \mathbb{R} \rightarrow \mathbb{R}$, taking one value and returning one value in the real numbers domain. In order to maintain consistency of representation we use the identity function, $id(x) = x$ to represent the cases where we do not apply any transformation.

As an example of this representation, consider a three variable regression problem with the variables set $x = \{x_1, x_2, x_3\}$ and the function corresponding to a linear regression of these variables $w_1 \cdot x_1 + w_2 \cdot x_2 + w_3 \cdot x_3$. Such expression would be represented as:

\begin{align}
T &= \{ t_1, t_2, t_3 \} \nonumber \\
t_1 &= ( [1,0,0], id ) \nonumber \\
t_2 &= ( [0,1,0], id ) \nonumber \\
t_3 &= ( [0,0,1], id ), 
\label{eq:linearterms}
\end{align}

\noindent and the linear regression $\hat{Y} = \mathbf{w} \cdot T$ is solved to determine the coefficients $\mathbf{w}$.

Similarly, the function $w_1 \cdot x_1^3 \cdot x_2 + w_2 \cdot \sin{(x_3)}$ would be represented as:

\begin{align}
T &= \{ t_1, t_2 \} \\
t_1 &= ( [3,1,0], id ) \\
t_2 &= ( [0,0,1], \sin ),
\end{align}

\noindent again solving the regression $\hat{Y} = \mathbf{w} \cdot T$ in order to obtain the original function.

Given a set of terms $T$ and a set of weights $W$ associated with each term, the computational complexity of evaluating the mathematical expression represented by $T$ for a given $d$-dimensional point $x$ can be determined by means of the number of terms $n$ and dimension $d$. Assuming a constant cost for calculating any transformation function or the power of a number, each term should evaluate $d$ power functions plus a transformation function, each term should also be multiplied by its corresponding weight. As such the computational complexity is $O(n \cdot (d + 2))$, or $O(n \cdot d)$.

\subsection{Symbolic Regression Search Tree}
\label{sec:alg}

The general idea of the proposed algorithm is to perform a tree-based search where the root node is simply a linear regression of the original variables and every child node is an expansion of the parent expression. The search is performed by expanding the tree in a breadth-first manner where every children of a given parent is expanded before exploring the next level of the tree.

The algorithm starts with a set of terms each of which corresponding to one of the variables without any interaction or transformation, as exemplified in Eq.~\ref{eq:linearterms}. This is labeled as the root node of our tree and this node is given a score calculated by:

\begin{equation}
score(model) = \frac{1}{1 + MAE(model)},
\label{eq:invmae}
\end{equation}

\noindent where $MAE(.)$ returns the mean absolute error of the linear regression model learned from the expression. After that, three different operators are applied to this expression: interaction, inverse interaction or transformation.

In the interaction operator, every combination of terms $(t_i, t_j)$ is enumerated creating the term $t_k = ( P_i + P_j, id )$, the addition operation of polynomials is simply the sum of the vectors $P_i$ and $P_j$. Likewise, the inverse interaction creates the term $t_k = ( P_i - P_k, id )$. Finally, the transformation operator creates new terms by changing the associated function of every $t_i$ with every function of a list of transformation functions.

Following the example given in Eq.~\ref{eq:linearterms}, the interaction operator would return the set $ \{ t_1 + t_1, t_1 + t_2, t_1 + t_3, t_2 + t_2, t_2 + t_3, t_3 + t_3 \}$ which would generate the new terms:

\begin{align*}
t_4 &= ( [2,0,0], id ) \\
t_5 &= ( [1,1,0], id ) \\
t_6 &= ( [1,0,1], id ) \\
t_7 &= ( [0,2,1], id ) \\
t_8 &= ( [0,1,1], id ) \\
t_9 &= ( [0,0,2], id ) 
\end{align*}

Given a set of $n_i$ terms in the $i$-th node of the search tree, this operation will have a computational complexity of $O(n_i^2)$

The inverse interaction operation would return the set $\{ t_1 - t_2, t_1 - t_3, t_2 - t_3, t_2 - t_1, t_3 - t_1, t_3 - t_2 \}$ and the new terms:

\begin{align*}
t_{10} &= ([1,-1,0], id ) \\
t_{11} &= ( [1,0,-1], id ) \\
t_{12} &= ( [0,1,-1], id ) \\
t_{13} &= ( [-1,1,0], id ) \\
t_{14} &= ( [-1,0,1], id ) \\
t_{15} &= ( [0,-1,1], id )  
\end{align*}

Similarly, this operation will have a computational complexity of $O(n_i^2)$

Finally, given the set of functions $\{ \sin, \log \}$ the transformation operator would return:

\begin{align*}
t_{16} &= ( [1,0,0], \sin ) \\
t_{17} &= ( [1,0,0], \log ) \\
t_{18} &= ( [0,1,0], \sin ) \\
t_{19} &= ( [0,1,0], \log ) \\
t_{20} &= ( [0,0,1], \sin ) \\
t_{21} &= ( [0,0,1], \log ) 
\end{align*}

Given a set of $m$ transformation functions and $n_i$ terms, this operation will have a computational complexity of $O(m \cdot n_i)$. The application of all these operations (i.e., each step of the algorithm) will then have a complexity of $O(m \cdot n_i + n_i^2)$.

After this procedure, every term that produces any indetermination (i.e., $\log{(0)}$) on the current data is discarded from the set of new terms. The score of each remaining term is then calculated by inserting the term into the parent node expression and calculating the model score with Eq.~\ref{eq:invmae}. Those terms that obtain a score smaller than the score of its parent are eliminated.

Finally, a greedy heuristic is performed to generate the child nodes. This heuristic expands the parent node expression by inserting each generated term sequentially and recalculating the score of the new expression, whenever the addition of a new term reduces the current score, the term is removed and inserted into a list of unused terms. After every term is tested, this new expression becomes a single child node and the process is repeated with the terms stored at the unused terms list, generating other child nodes. This is repeated until the unused terms list is empty. Notice that because only the terms that improved upon the parent expression are used in this step, every term will eventually be used in one of the child nodes.

After every child node is created, the corresponding expressions are simplified by eliminating every term that has an associated coefficient (after applying a linear regression) smaller than a given threshold $\tau$ set by the user. This whole procedure is then repeated, expanding each of the current leaf nodes, until a stop criteria is met.

Notice that the greedy heuristic prevents the enumeration of every possible expression, thus avoiding unnecessary computation. But, on the other hand, it may also prevent the algorithm from reaching the optimum expression. The elimination of terms that have an importance below $\tau$ also helps avoiding an exponential growth of the new terms during the subsequent application of the operators. 

In the worst case scenario, when every new term increases the score of the expression and has a coefficient higher than $\tau$ in every iteration, the number of terms of an expression will still increase exponentially. In this scenario, the expansion will have a number of terms proportional to $n_i^2$. After $k$ iterations, the expected number of terms will be proportional to $n_0^{2k}$, with $n_0$ being the number of original variables of the data set.

Another implementation procedure devised to avoid an exponential growth is the use of the parameters $minI$ and $minT$ to control at what depth of the search tree the Inverse and Transformation operators will start to be applied. With these parameters, the expression may first expand to a set of terms containing only positive polynomial interaction with the expectation to have enough information to be simplified when applying the other operators.

Due to the tree-search nature, the algorithm will be named as Symbolic Regression Tree, or \textit{SymTree} for short, the pseudo-algorithm is illustrated in Alg.~\ref{alg:symtree} together with the auxiliary functions in Algs.~\ref{alg:expand}~and~\ref{alg:greedy}.

\begin{algorithm}[ht!]
  \SetKwData{Root}{root}\SetKwData{Leaves}{leaves}\SetKwData{Leaf}{leaf}\SetKwData{Nodes}{nodes}
  \SetKwFunction{LinearExpression}{LinearExpression}\SetKwFunction{Expand}{Expand}\SetKwFunction{Score}{Score}
  \SetKwInOut{Input}{input}\SetKwInOut{Output}{output}
  \Input{data points $X$ and corresponding set of target variable $Y$, simplification threshold $\tau$, minimum iteration for inverse and transformation operators $minI$, $minT$.}
\Output{symbolic function $f$}
\BlankLine

\tcc{create root node (Eq.~\ref{eq:linearterms}).}
\Root $\leftarrow$ \LinearExpression{$X$}\;
\Leaves $\leftarrow$ \{\Root\}\;

\While{criteria not met}{
    \Nodes $\leftarrow \varnothing$\;
    \For{\Leaf $in$ \Leaves}{
        \Nodes $\leftarrow$  \Nodes $\cup$ \Expand{\Leaf, $\tau$, $it > minI$, $it > minT$}\;
    }
    \Leaves $\leftarrow$ \Nodes\;

}
\Return $arg \max $ \{\Score(\Leaf) for \Leaf $\in$ \Leaves\}\;
\caption{SymTree algorithm}
\label{alg:symtree}
\end{algorithm}

The Alg.~\ref{alg:symtree} is a straightforward and abstract description of the SymTree algorithm. The \textit{Expand} function described in Alg.~\ref{alg:expand} gives further detail of the inner procedure of node expansion. In this function the Interaction, Inverse and Transformation operators are applied, followed by the \textit{GreedySearch} function responsible for the creation of a new expanded expression (Alg.~\ref{alg:greedy}). The \textit{Simplify} function removes the terms with a coefficient smaller than $\tau$ of the corresponding linear regression.

\begin{algorithm}[ht!]
  \SetKwData{Terms}{terms}\SetKwData{Node}{node}\SetKwData{Children}{children}\SetKwData{Term}{term}\SetKwData{NewNode}{new\_node}
  \SetKwFunction{Interaction}{Interaction}\SetKwFunction{Inverse}{Inverse}\SetKwFunction{Transformation}{Transformation}
  \SetKwFunction{Score}{Score}\SetKwFunction{GreedySearch}{GreedySearch}\SetKwFunction{Simplify}{Simplify}
  \SetKwInOut{Input}{input}\SetKwInOut{Output}{output}
  \Input{expression $node$ to be expanded, simplification threshold $\tau$, booleans indicating whether to apply the Inverse ($inv$) and the Transformation ($trans$) operators.}
\Output{set of child nodes $children$.}
\BlankLine

\tcc{create candidate terms.}
\Terms $\leftarrow$ \Interaction{\Node} $\cup$ \Inverse{\Node, $inv$} $\cup$ \Transformation{\Node, $trans$}\;
\Terms $\leftarrow$ [\Term $\in$ \Terms if \Score{\Node $+$ \Term} $>$ \Score{\Node}]\;

\tcc{generate nodes.}
\Children $\leftarrow \varnothing$\;
\While{\Terms $\neq \varnothing$}{
    \NewNode, \Terms $\leftarrow$ \GreedySearch{\Node, \Terms}\;
    \NewNode $\leftarrow$ \Simplify{\NewNode, $\tau$}\;
    \Children $\leftarrow$ \Children $\cup$ \NewNode\;
}

\tcc{guarantees that it returns at least one child.}
\If{\Children $\neq \varnothing$}{
    \Return \Children\;
}\Else{
    
    \Return \{\Node\}\;
}
\caption{Expand function}
\label{alg:expand}
\end{algorithm}

\begin{algorithm}[ht!]
  \SetKwData{Terms}{terms}\SetKwData{Node}{node}\SetKwData{Term}{term}
  \SetKwFunction{Score}{Score}
  \SetKwInOut{Input}{input}\SetKwInOut{Output}{output}
  \Input{expression $node$ to be expanded and the list of candidate $terms$.}
\Output{the expanded $node$ and the unused $terms$.}
\BlankLine

\tcc{expand terms.}
\For{\Term $\in$ \Terms}{
    \If{\Score{\Node $+$ \Term} $>$ \Score{\Node}}{
        \Node $\leftarrow$ \Node $+$ \Term\;
    }
}
\Terms $\leftarrow$ [\Term $\in$ \Terms if \Term $\notin$ \Node]\;
\Return \Node, \Terms\;

\caption{GreedySearch function}
\label{alg:greedy}
\end{algorithm}

\begin{figure}[ht!]
\centering
\subfigure[]{
\includegraphics[height=0.2\textheight]{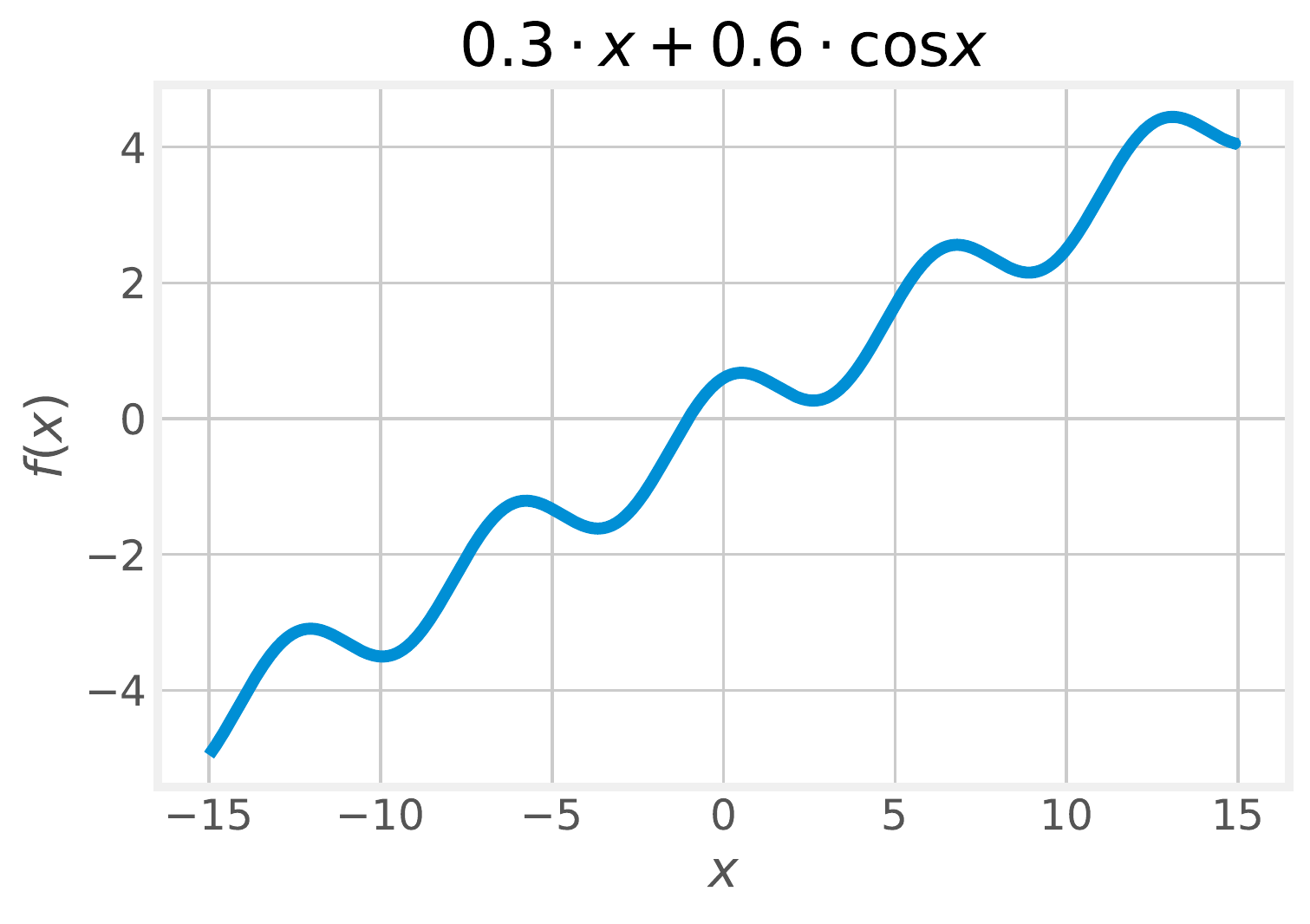}
\label{fig:expr1}
}
\subfigure[]{
\includegraphics[height=0.2\textheight]{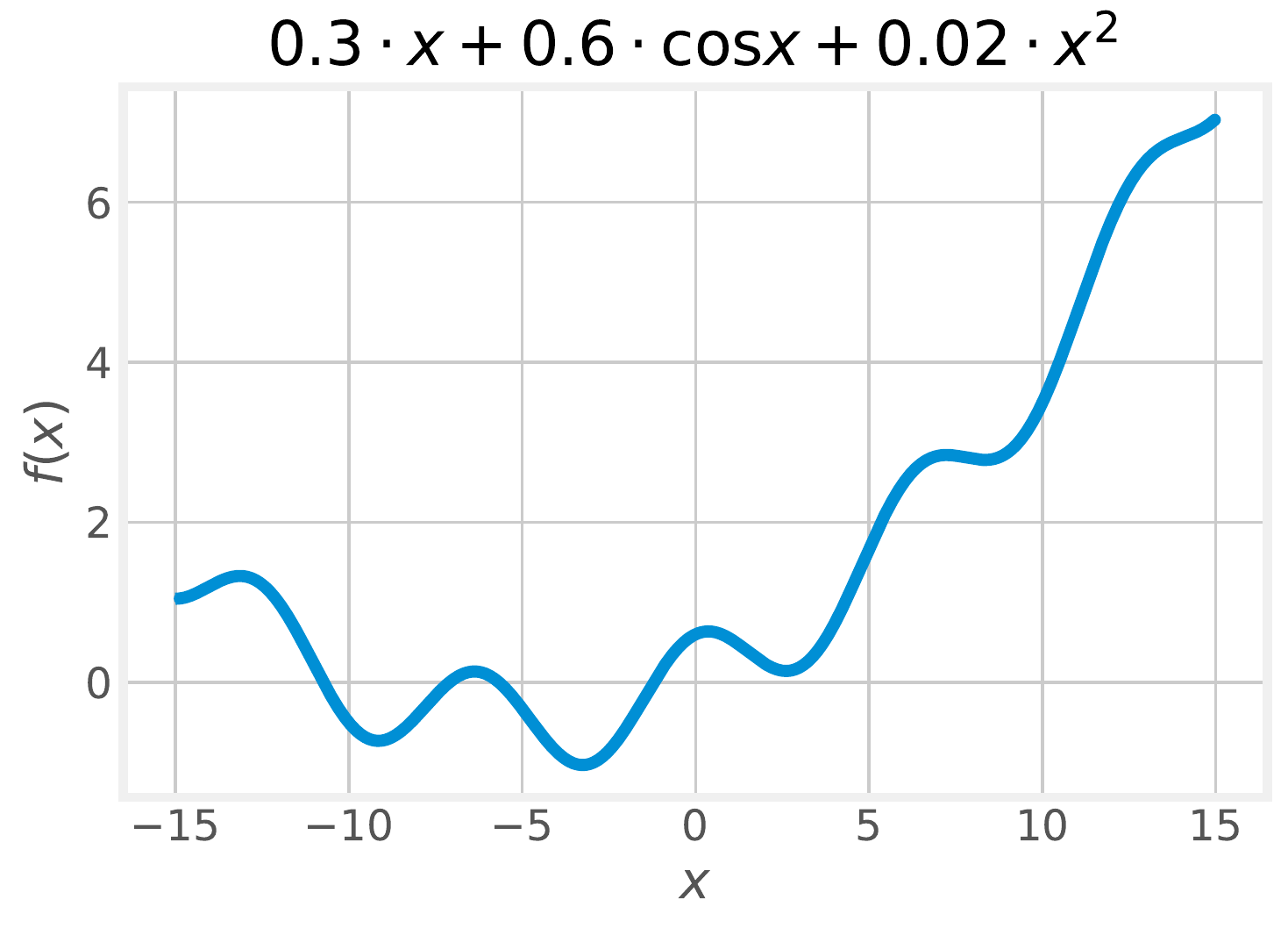}
\label{fig:expr2}
}
\caption{Example of the transition between a parent expression $0.3x + 0.6\cos{(x)}$ (a) and the child expression $0.2x + 0.6\cos{(x)} + 0.02x^2$ (b) with SymTree.}
\label{fig:expr}
\end{figure}

\begin{figure}[ht!]
\centering
\includegraphics[height=0.25\textheight]{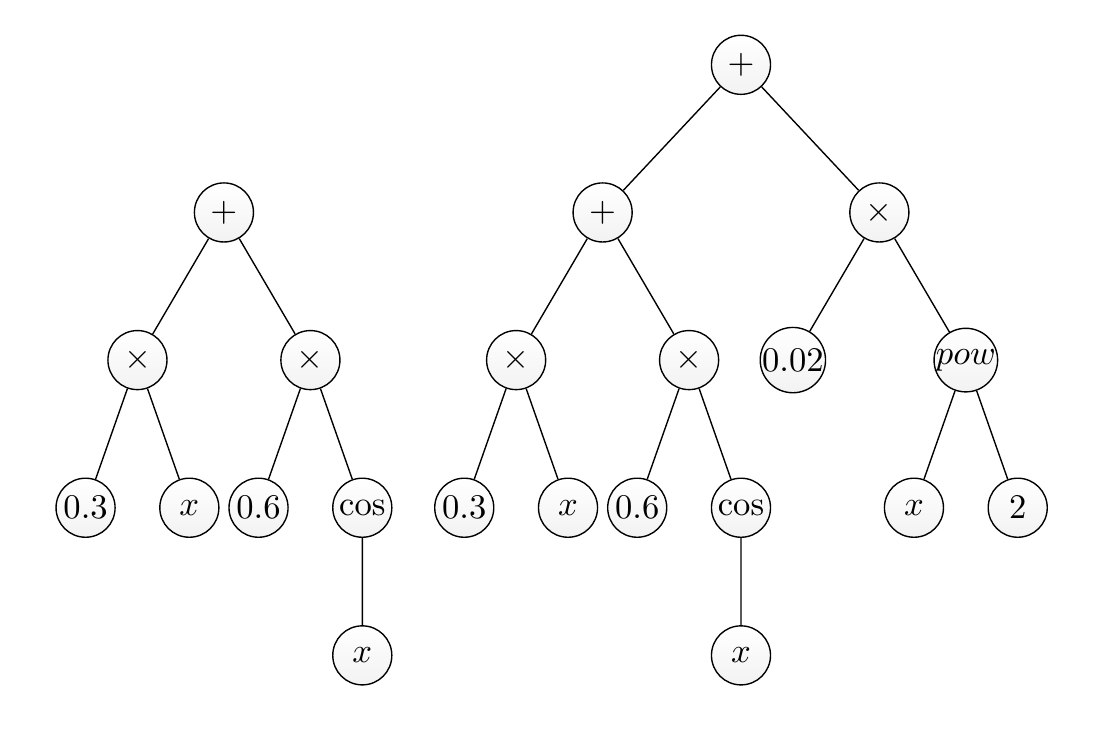}
\caption{Corresponding expression trees with standard GP representation of expressions depicted in Fig.~\ref{fig:expr}.}
\label{fig:gpexpr}
\end{figure}

The application of a linear regression helps to regulate the amount of change the expression will undertake from one node to the other. This allows a smooth transition from the previous solution to the next if required. As an example, suppose that the current node contains the expression $0.3x + 0.6\cos{(x)}$ (Fig.~\ref{fig:expr1}) and within one of its child the interaction $x^2$ is inserted. After solving the linear regression for the new expression the coefficients may become $0.2x + 0.6\cos{(x)} + 0.02x^2$ (Fig.~\ref{fig:expr2}). Notice that in order to achieve the same transition from $0.3x + 0.6\cos{(x)}$ to $0.2x + 0.6\cos{(x)} + 0.02x^2$ in GP, the expression $x^2$ should be a part of another expression tree and both solutions should be combined through crossover at the exact point as illustrated in Fig.~\ref{fig:gpexpr}.

\section{Experimental Results}
\label{sec:experiments}

The goal of the IT data structure and SymTree algorithm, proposed in this paper is to achieve a concise and descriptive approximation function that minimizes the error to the measured data. As such, not only we should minimize an error metric (i.e., absolute or squared error) but we should also minimize the size of the final expression.

So, in order to test both SymTree and the representation, we have performed experiments with a total of $17$ different benchmark functions commonly used in the literature, extracted from~\cite{karaboga2012artificial, trujillo2016neat, kattan2016gp, kattan2015surrogate}. 

These functions are depicted in Table~\ref{tab:benchmark} with the information whether the function itself is expressible by the IT data structure or not. Notice that even though some functions are not expressible, the algorithm can still find an approximation of such functions. In total, from the $17$ functions, $8$ are not expressible, or barely half of the benchmark. As a comparison, for the EFS only $5$ of these functions are not expressible, while the remaining GP algorithms tested here can represent each one of these benchmark functions.

For every benchmark function, $600$ samples were randomly generated with uniform distribution within the range $[-5, 5]$, with the exception of $F7$ that was sampled within $[0,2]$ in order to avoid indefinitions. These samples were split in half as training and testing data sets. All the tested algorithms had access only to the training data during the model fitting step and, afterwards, the mean absolute error of the fitted model was calculated using the testing data.

\begin{table}[ht!]
\centering
\caption{Benchmark functions used for the comparative experiments.}
\begin{tabular}{c|c|c}
\hline
\textbf{Function} & \textbf{expressible} \\
\hline\hline
$F_1 = x^3 + x^2 + 5*x$ & Y\\
$F_2 = x^4 + x^3 + x^2 + x$ & Y\\
$F_3 = x^5 + x^4 + x^3 + x^2 + x$  & Y\\
$F_4 = x^6 + x^5 + x^4 + x^3 + x^2 + x$ & Y\\
$F_5 = sin(x^2)cos(x) - 1$ & N\\
$F_6 = sin(x) + sin(x + x^2)$ & N\\
$F_7 = log(x+1) + log(x^2+1)$  & Y\\
$F_8 = 5*\sqrt{\|x\|}$ & Y\\
$F_9 = sin(x) + sin(y^2)$ & Y\\
$F_{10} = 6sin(x)cos(y)$  & N\\
$F_{11} = 2 - 2.1cos(9.8x)sin(1.3w)$ & N\\
$F_{12} = \frac{e^{-(x-1)^2}}{1.2 + (y-2.5)^2}$ & N\\
$F_{13} = \frac{10}{5 + \sum_{i=1..5}{(x_i-3)^2}}$  & N\\
$F_{14} = x_1x_2x_3x_4x_5$  & Y\\
$F_{15} = \frac{x^6}{x^3 + x^2 + 1}$ & N \\
$F_{16} = \frac{x}{1 - \log(x^2 + x + 1)}$ & N \\
$F_{17} = 100 + \log(x^2) + 5\sqrt(|x|)$ & Y \\
\hline\hline
\end{tabular}
\label{tab:benchmark}
\end{table}

The proposed approach was compared against other Symbolic and traditional Regression algorithms. The representative set of traditional regression algorithms was chosen to comprehend from the simpler to more complex models. As such this set is composed by a Linear Regression~\cite{rao2009linear}, Linear Regression with Polynomial Features~\cite{scholkopf2001statistical} and Gradient Tree Boosting~\cite{chen2015xgboost}. All of these regression models are provided by the \emph{scikit-learn} Python package~\cite{pedregosa2011scikit}.

As for the Symbolic Regression algorithms set, the choices were the ones already described in Sec.~\ref{sec:review}: \emph{neat}-GP, EFS, sGP and SSGP, for the first two we have used the provided source code in Python~\footnote{https://github.com/saarahy/neatGP-deap} and in Java~\footnote{https://github.com/flexgp/efs}, respectively, and for the last two we have used the reported values from the literature.

The algorithm parameters was set by applying a grid search within a set of pre-defined parameter values, each combination of parameters within these set was tested using only the training data and the, best model obtained by each algorithm was used to measure the goodness-of-fit to the testing data. Next, each algorithm will be briefly explained together with the parameters set used for the grid search.

The Linear Regression (LR) uses the Ordinary Least Square in order to estimate its parameters. Besides testing this model, we also transformed the data by generating Polynomial Features (PF) in order to allow the modeling of non-linear relationship as linear. We have tested polynomials of degrees in the range $[2, 6]$. 

Finally, within the traditional algorithms, the Gradient Tree Boosting(GB) creates the regression model by means of a boosting ensemble outperforming many regression algorithms~\cite{chen2016xgboost, chen2015higgs}. This algorithm has two main parameters that improves the performance, but increases the size of the final model: the number of boosting stages, improving the robustness, and the maximum depth of each tree, minimizing the error of each regressor. The number of boosting stages was tested within the set $\{100, 500\}$ after verifying that a larger number of estimators would just increase the length of the final expression without significantly reducing the regression error on the tested functions. The maximum depth was tested in the range of $[2, 6]$ following the same rationale.

Within the Symbolic Regression algorithms, EFS only allows one parameter to be set, the maximum time in minutes allowed to search for a solution. As such we allowed a total amount of $1$ minutes, more than what was used by every other algorithm within this benchmark. The \emph{neat}-GP algorithm was tested with a population size of $500$, evolved through $100$ iterations. The crossover rate was tested with the values $\{0.5, 0.7, 0.9\}$, mutation rate with $\{0.1, 0.3, 0.5\}$, survival threshold with $\{0.3, 0.4, 0.5\}$, specie threshold value with $\{0.1, 0.15, 0.2\}$ and $\alpha = 0.5$, all values fixed values as suggested by the authors. The other two GP algorithms, sGP and SSGP, had neither of their parameters adjusted for this test since we will only use the reported values in their respective papers.

Regarding the SymTree algorithm, the threshold parameter $\tau$ was tested within the range $\{1e-6, 1e-5, 1e-4, 1e-3, 1e-2\}$, the $minI$ parameter that controls the iteration that it starts to apply the inverse interation operator was tested within $[1, 10[$, the $minT$ parameter, that controls at which iteration it starts to apply the transformation functions was tested within the range $[5, 10[$ and the total number of iterations steps performed by the algorithm was set as $minI + minT + it$ with $it$ tested within the range $[0,5]$. Notice that despite testing only a small number of iterations, it is worth noticing that after $n$ iterations the algorithm can represent polynomials of degree $2^n$ only using the interaction operator.

The functions set used for the transformation functions for the Symbolic Regression algorithms were fixed to the set $\{ \sin(x), \cos(x), \tan(x), \sqrt{\|x\|}, \log(x), \log(x+1) \}$. This set was chosen in order to allow the IT data structure to correctly represent most of the functions tested here.

\subsection{Accuracy comparison}
\label{sec:mae}

Our first analysis will be how well SymTree fared against the contenders with respect to the goodness of fit for each data set from the benchmark. In this analysis we are only concerned about the minimization of an error function. For this purpose, we have measured the Mean Absolute Error to allow a proper comparison with the reported results in~{kattan2016gp,kattan2015surrogate}.

The experiments with the non-deterministic algorithms (GB, \emph{neat}-GP, EFS) were repeated $30$ times and the average of these values are reported in the following tables and plots. Since SymTree is deterministic we have performed a One-sample Wilcoxon test comparing the result obtained by SymTree against the results obtained by the best or second best (whenever SymTree had the best result) algorithm with the difference being considered significant with a p-value $< 0.05$.

In Table~\ref{tab:maeResults} we can see the comparative results summarized by the Mean Absolute Error. From this table we can see that, as expected, SymTree was capable of finding the optimal solution for every function that the IT data structure was capable of expressing. In $5$ of these functions the proposed algorithm could find a significantly better solution than every other tested algorithm and, in $13$ of these functions it also found a significantly better solution than every other Symbolic Regression algorithm. Notice that in $9$ of the benchmark functions it achieved the optimal solution, sometimes drawing with Polynomial Features and/or Gradient Boosting. Even when it did not achieve the best solution it still managed to find a competitive approximation.

It is interesting to notice that \emph{neat}-GP seems to have a difficulty when dealing with polynomial functions, probabily due to its bloat control that penalizes tree with higher heights, favoring the trigonometric approximations (the generated expressions extensively used the trigonometric functions). The surrogate models were competitive, though never finding the best solution, in almost every one of their reported values, but this should still be validated using the original source code. The EFS algorithm performed competitively against SymTree in many functions but had some difficulties dealing with polynomial functions, similar to \emph{neat}-GP, probably due to the use of Lasso penalty function that tries to favor smaller expressions.

From the traditional algorithms, surprisingly enough Linear Regression obtained some very competitive results even finding better approximation than every other contender in $2$ different occasions. The use of polynomial features improved the Linear Regression capabilities even further. It is important to notice that, even though it should be expected that PF would find a perfect fit for $F4$, the fourth and sixth degree polynomials are very close to each other, as such the sampled data for the training set can have deceived this (and the others) approach to fit the wrong model. Even though this did not happen with SymTree, it cannot be assured that it could have happend with different samples. Finally, GB behaved more consistently close to the best approximation with just two exceptions. $F4$ and $F14$, achieving the best approximation in $7$ benchmark functions (effectively being the solo winner in $5$ of them).


\begin{table}[ht!]
\centering
\caption{Mean Absolute Error obtained by each algorithm on the benchmark functions. The results in which SymTree results are better when compared to all the other algorithms are marked in bold, those results in which SymTree performed better when compared to all the other Symbolic Regression algorithms are marked in italic.}
\begin{tabular}{ccccccccc}
\hline\hline
algorithm &      LR &     PF &    GB &    \emph{neat}-GP  &  SSGP & sGP & EFS &  SymTree \\
Function &         &        &       &         &       &      \\
\hline
F1       &   17.48 &   0.00 &  0.49 &  108.91 &  -- & 1.10 & 6.43 & $\mathit{0.00}$ \\
F2       &  131.99 &   0.00 &  1.83 &  284.70 & -- & -- & 6.62 & $\mathit{0.00}$ \\
F3       &  483.36 &   0.00 &  8.52 & 1025.46 & -- & -- & 32.50 & $\mathit{0.00}$ \\
F4       & 2823.94 & 247.72 & 58.39 &  813.25 & -- & -- & 47.55 & $\mathbf{0.00}$ \\
F5       &    0.32 &   0.30 &  0.03 &    0.65 & -- & -- & 0.25 & 0.23 \\
F6       &    0.79 &   0.60 &  0.07 &    1.95 & -- & -- & 0.57 & 0.58 \\
F7       &    0.02 &   0.00 &  0.00 &    0.58 & -- & -- & 0.01 & $\mathit{0.00}$ \\
F8       &    2.22 &   0.56 &  0.03 &    0.95 & -- & 0.99 & 0.14 & $\mathbf{0.00}$ \\
F9       &    0.78 &   0.58 &  0.18 &    4.45 & -- & -- & 0.24 & $\mathbf{0.00}$ \\
F10      &    2.26 &   2.05 &  0.95 &    8.74 & 2.31 & -- & 1.88 & $\mathbf{0.72}$ \\
F11      &    0.81 &   0.83 &  0.86 &    2.91 &  -- & -- & 0.85 & 0.82 \\
F12      &    0.06 &   0.06 &  0.01 &    2.00 &  0.07 & -- & 0.04 & 0.06 \\
F13      &    0.06 &   0.04 &  0.03 &    1.41 &  0.14 & -- & 0.05 & $\mathit{0.03}$ \\
F14      &   61.26 &   0.00 & 71.87 &    3.62 & 76.08 & -- & 90.50 & $\mathit{0.00}$ \\
F15      &   22.41 &   8.63 &  7.00 &  188.11 & -- & 13.45 & 12.57 & $\mathit{8.88}$ \\
F16      &    5.21 &   5.82 &  5.50 &    6.26 &  -- & 8.89 & 6.71 & $\mathit{6.31}$ \\
F17      &    3.88 &   1.34 &  0.11 &    3.78 &  -- & 5.90 & 0.26 & $\mathbf{0.00}$ \\
\hline\hline
\end{tabular}
\label{tab:maeResults}
\end{table}

Regarding the execution time, among the tested algorithms (with the exception of sGP and SSGP, that was not tested on the same machine) took roughly the same time. The Linear Regression and Polynomial Features took less than a second, on average, to fit each benchmark function. Gradient Boosting took $2$ seconds on average while SymTree took an average of $6$ seconds. Finally, \emph{neat}-GP and EFS took $30$ and $60$ seconds, respectively. All of these algorithms were implemented in Python 3~\footnote{notice that Linear Regression, Polynomial Features and Gradient Boosting may have additional optimizations unknown to this author.}, with the exception of EFS that was implemented in Java. All of these experiments were run at a Intel Core i5, 1.7GHz with 8GB of RAM under Debian Jessie operational system.


\subsection{Compromise between accuracy and simplicity}
\label{sec:compromise}

\begin{figure}[ht!]
\centering
  \subfigure[]{
  \includegraphics[trim = 0mm 0mm 30mm 5mm, clip, width=0.31\textwidth]{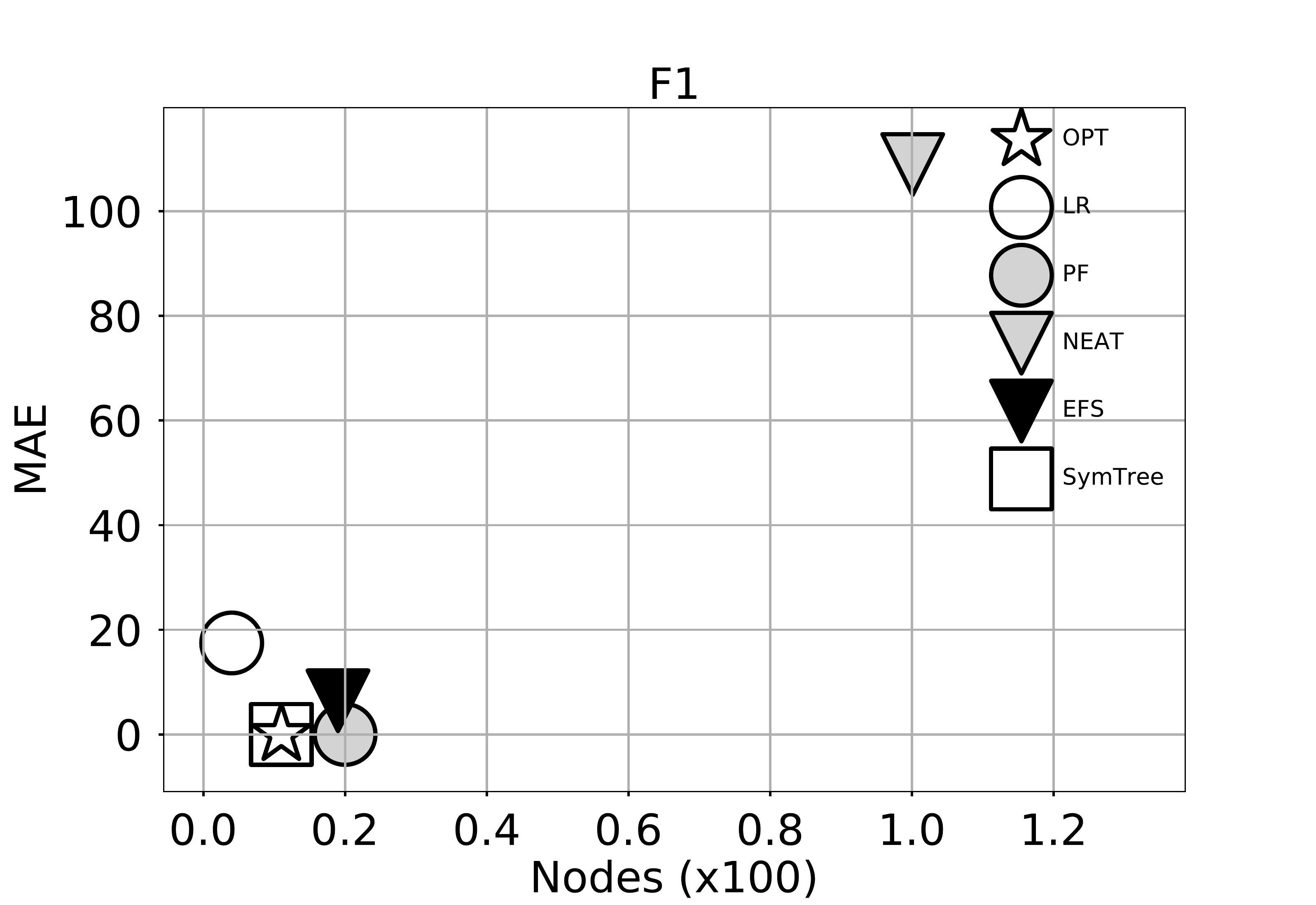}
  }
  \subfigure[]{
  \includegraphics[trim = 0mm 0mm 30mm 5mm, clip, width=0.31\textwidth]{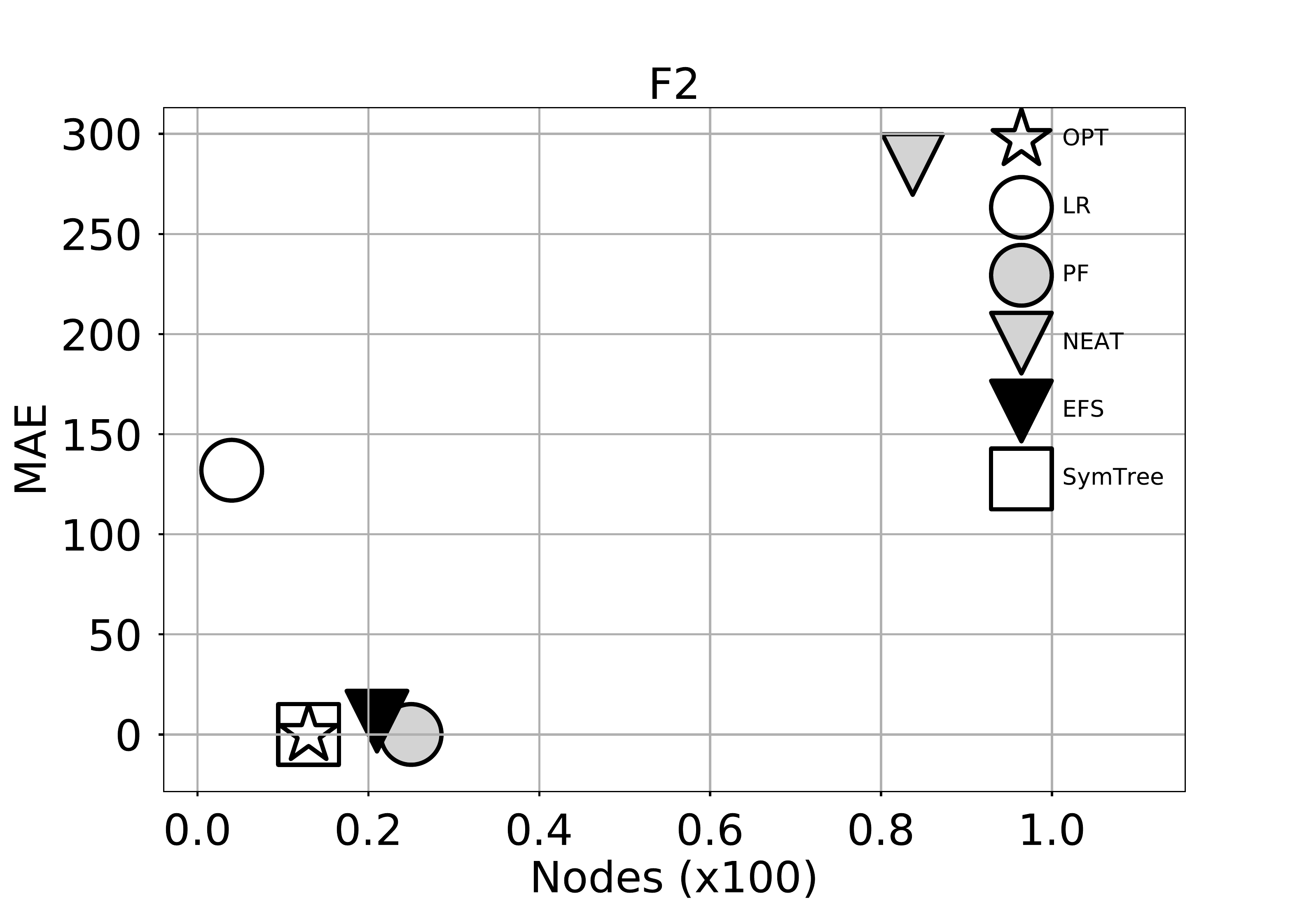}
  }
  \subfigure[]{
  \includegraphics[trim = 0mm 0mm 30mm 5mm, clip, width=0.31\textwidth]{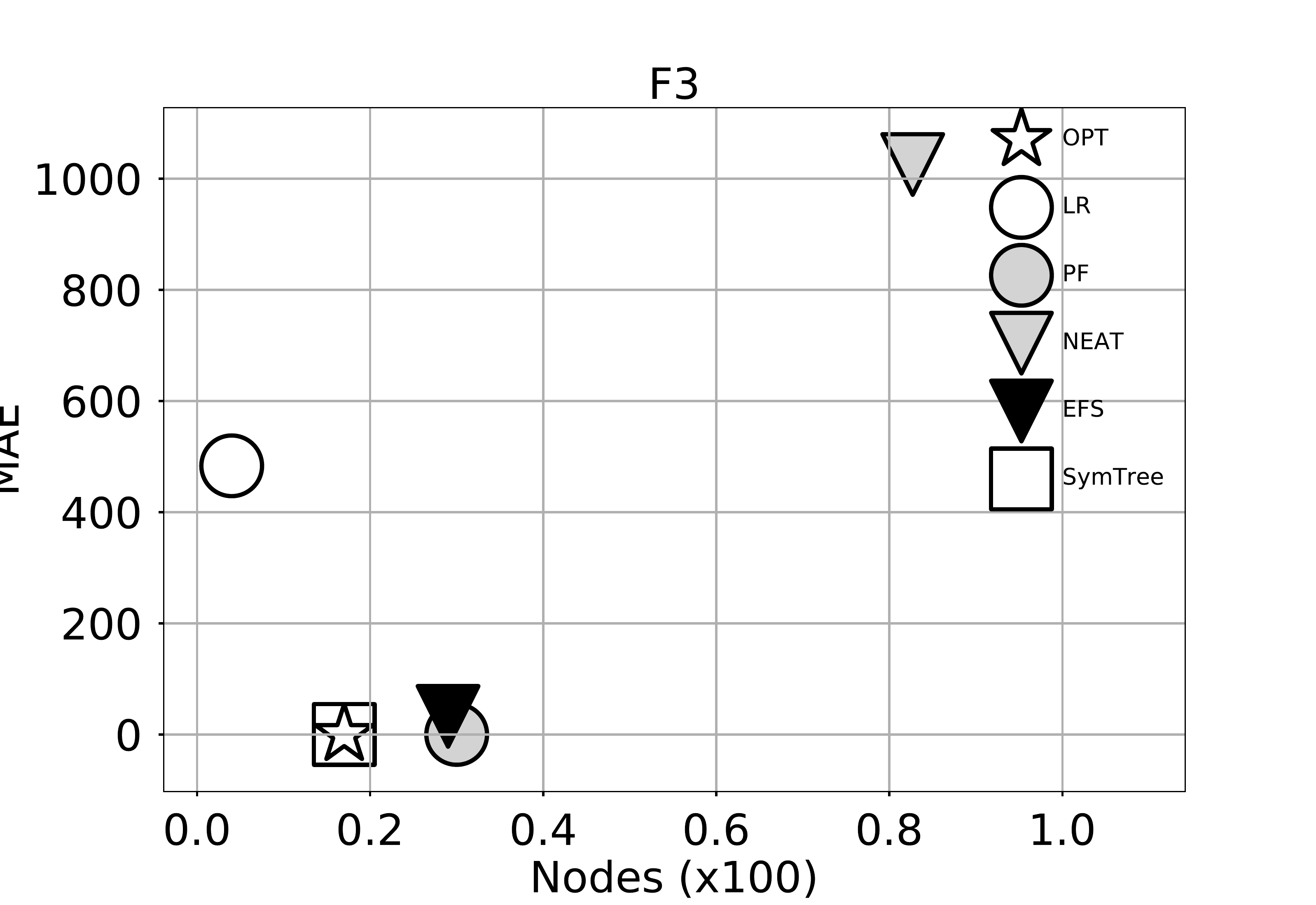}
  }
  \subfigure[]{
  \includegraphics[trim = 0mm 0mm 30mm 5mm, clip, width=0.31\textwidth]{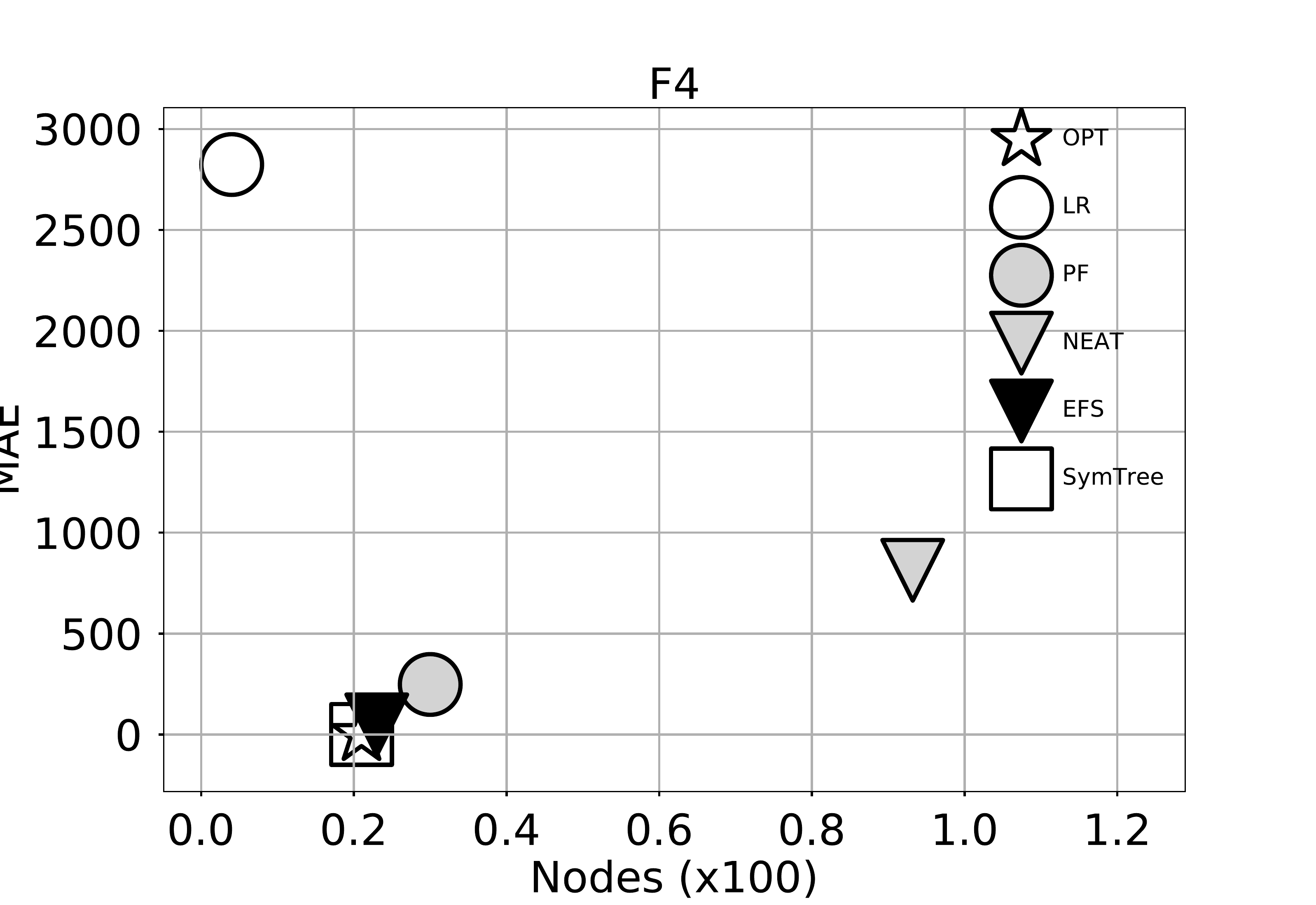}
  }
  \subfigure[]{
  \includegraphics[trim = 0mm 0mm 30mm 5mm, clip, width=0.31\textwidth]{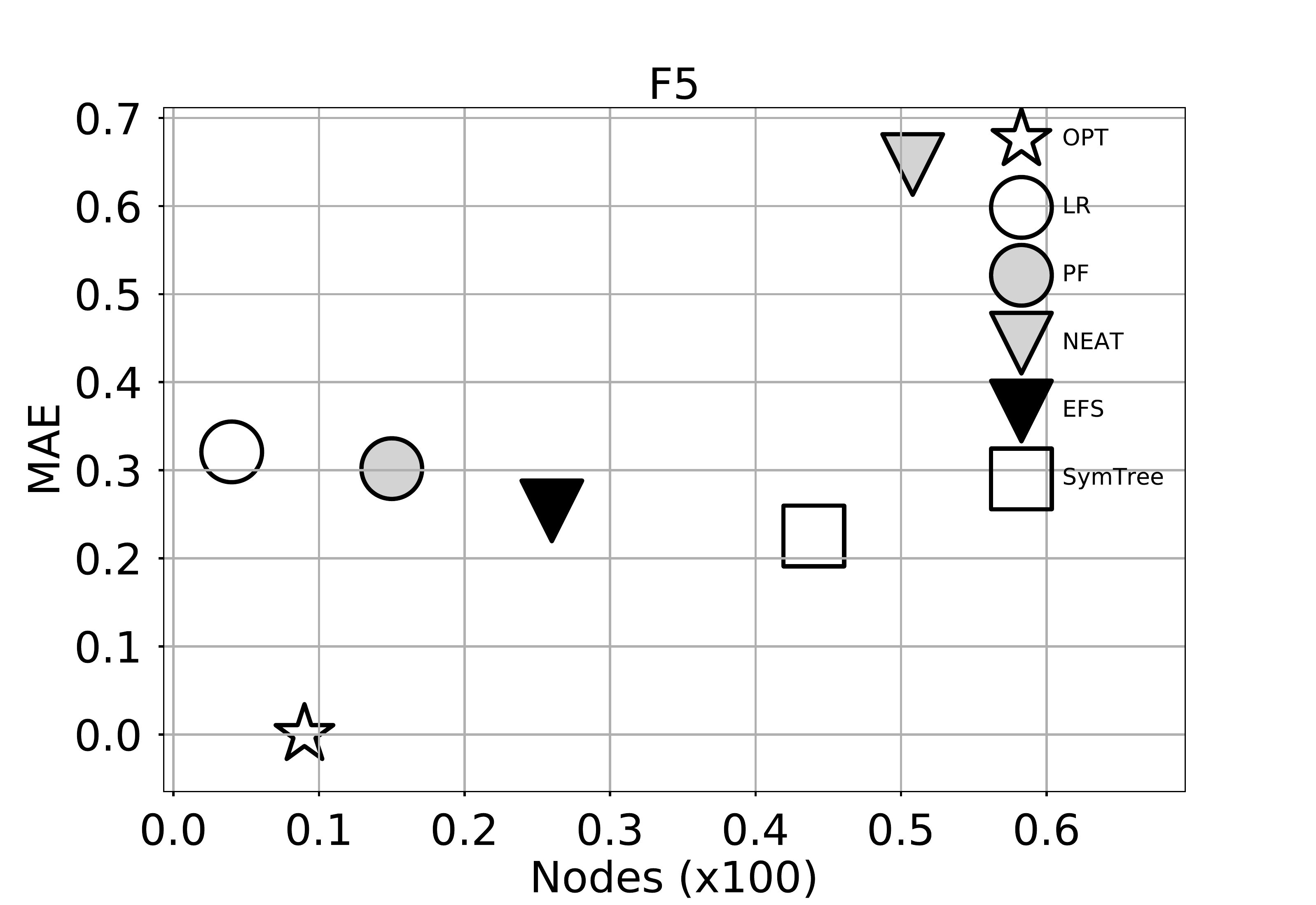}
  }
  \subfigure[]{
  \includegraphics[trim = 0mm 0mm 30mm 5mm, clip, width=0.31\textwidth]{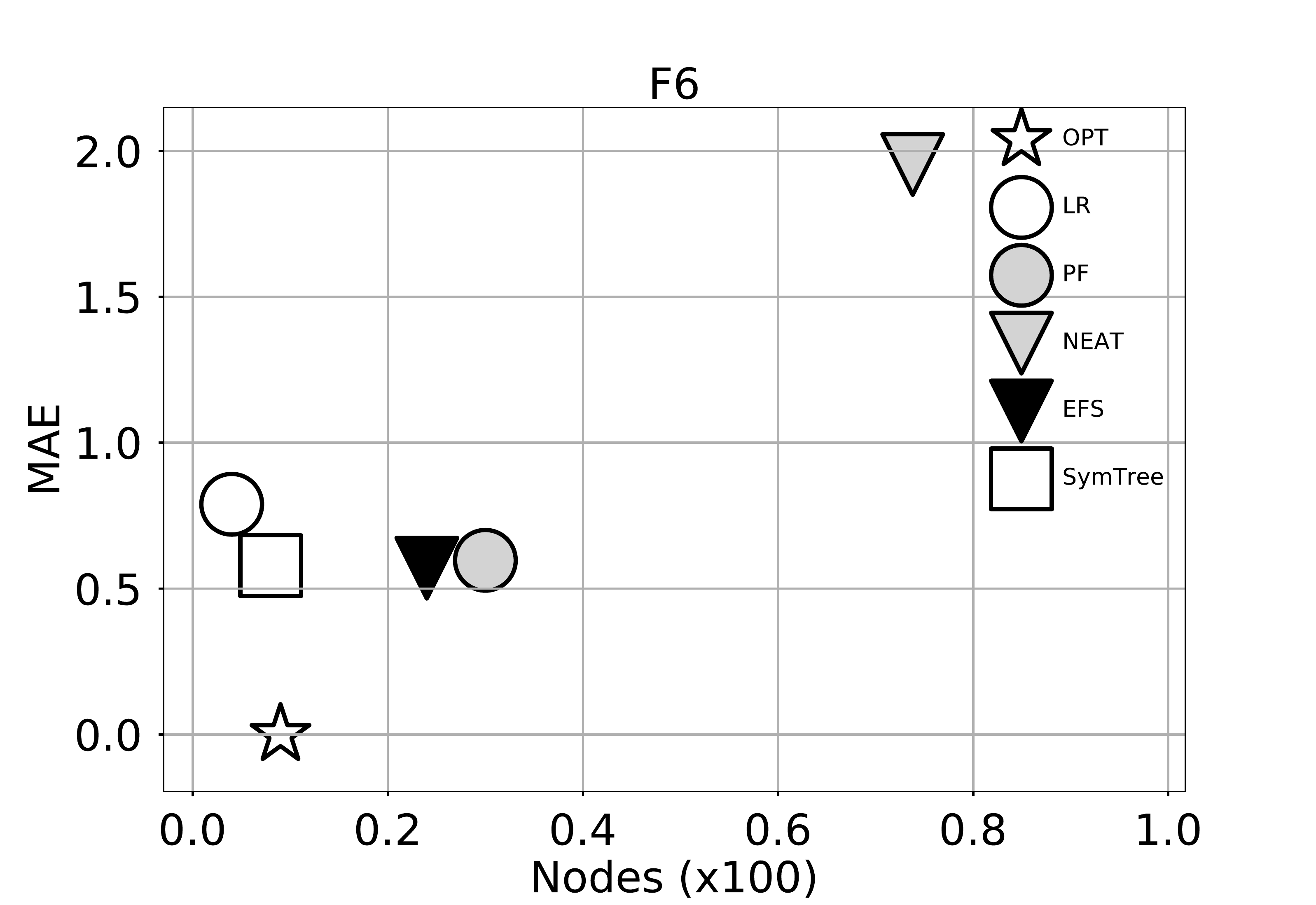}
  }

\caption{Compromise between accuracy and simplicity for functions F1 to F6.}
\label{fig:compromise1}
\end{figure}

\begin{figure}[ht!]
\centering
  \subfigure[]{
  \includegraphics[trim = 0mm 0mm 30mm 5mm, clip, width=0.31\textwidth]{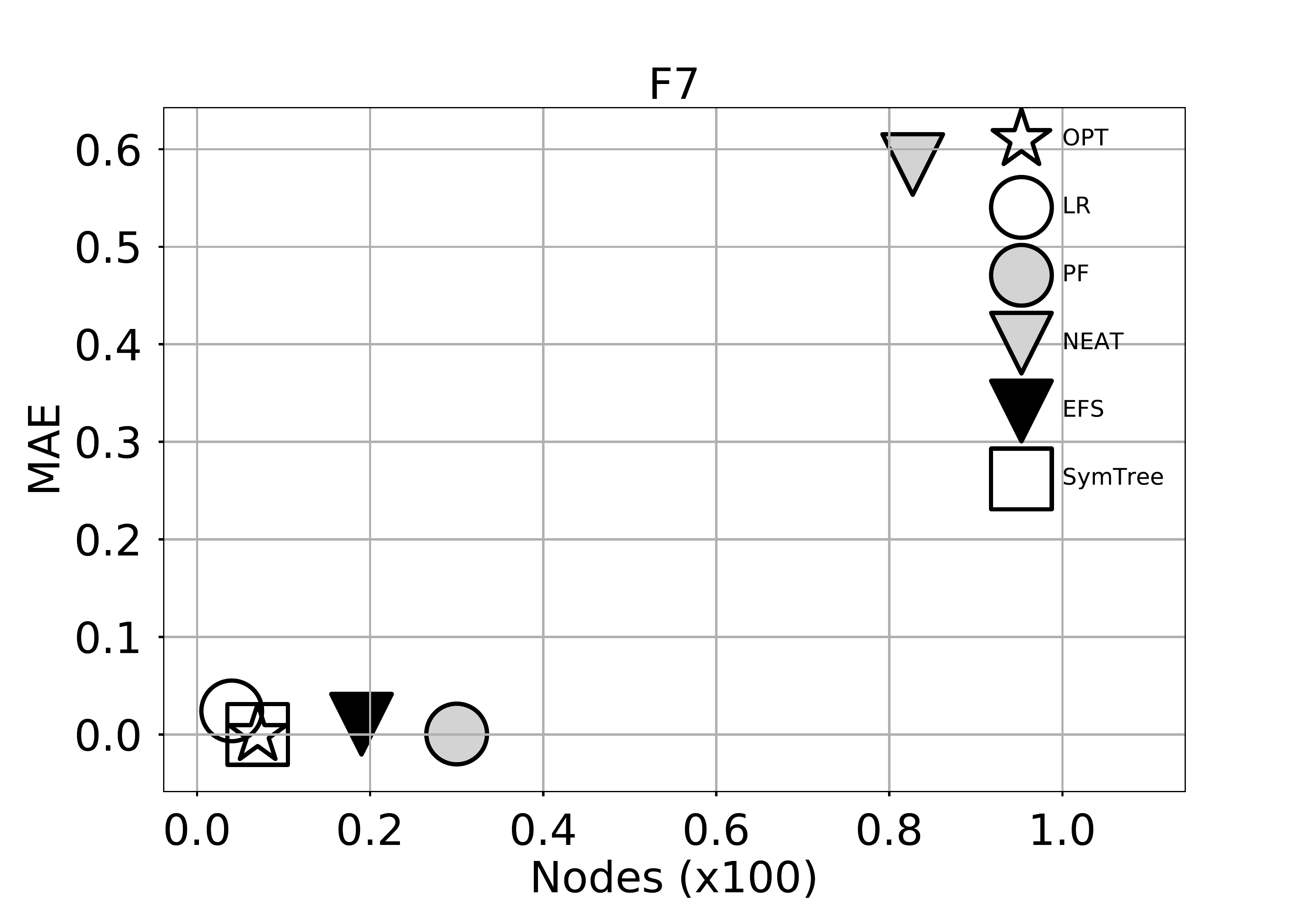}
  }
  \subfigure[]{
  \includegraphics[trim = 0mm 0mm 30mm 5mm, clip, width=0.31\textwidth]{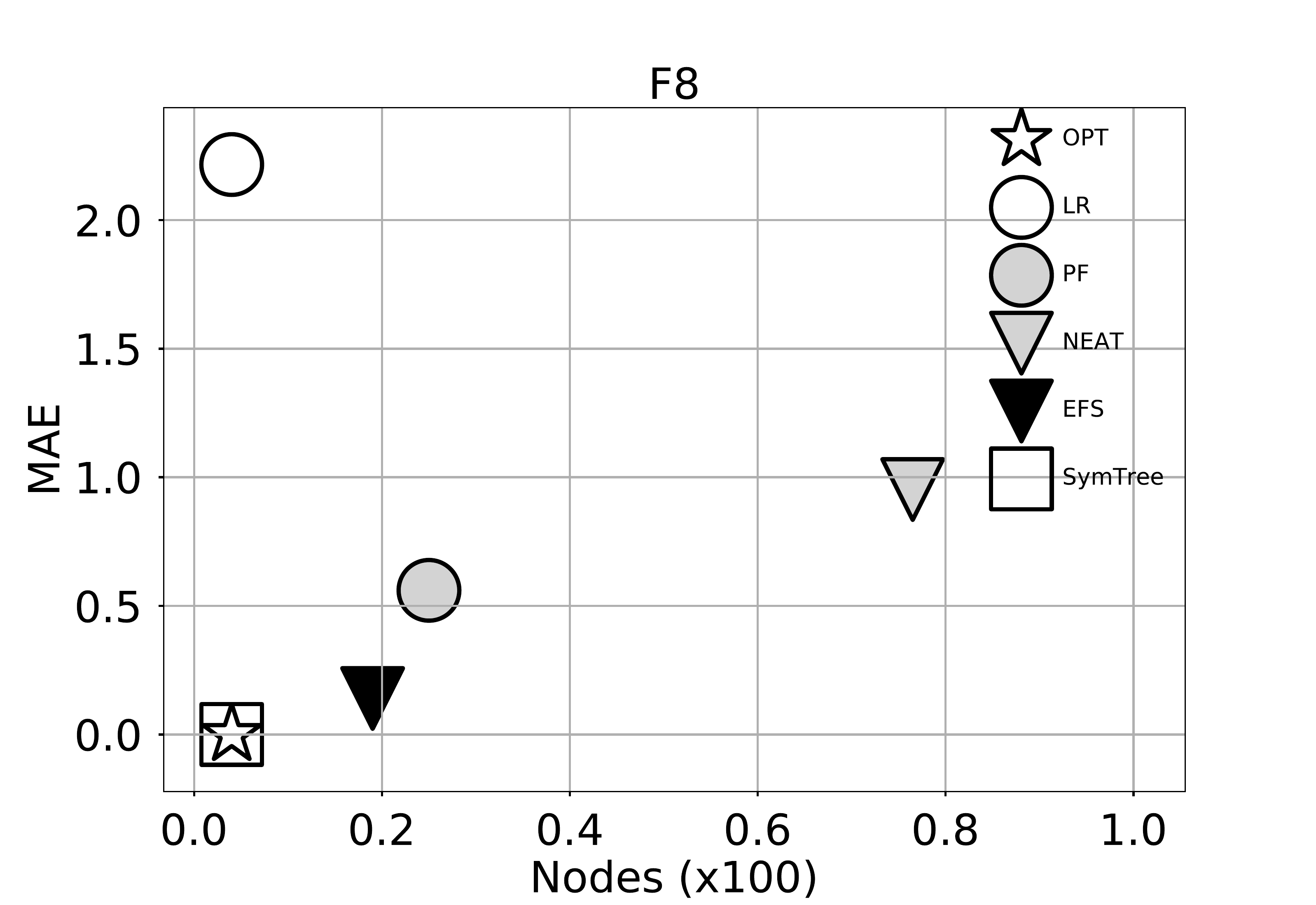}
  }
  \subfigure[]{
  \includegraphics[trim = 0mm 0mm 30mm 5mm, clip, width=0.31\textwidth]{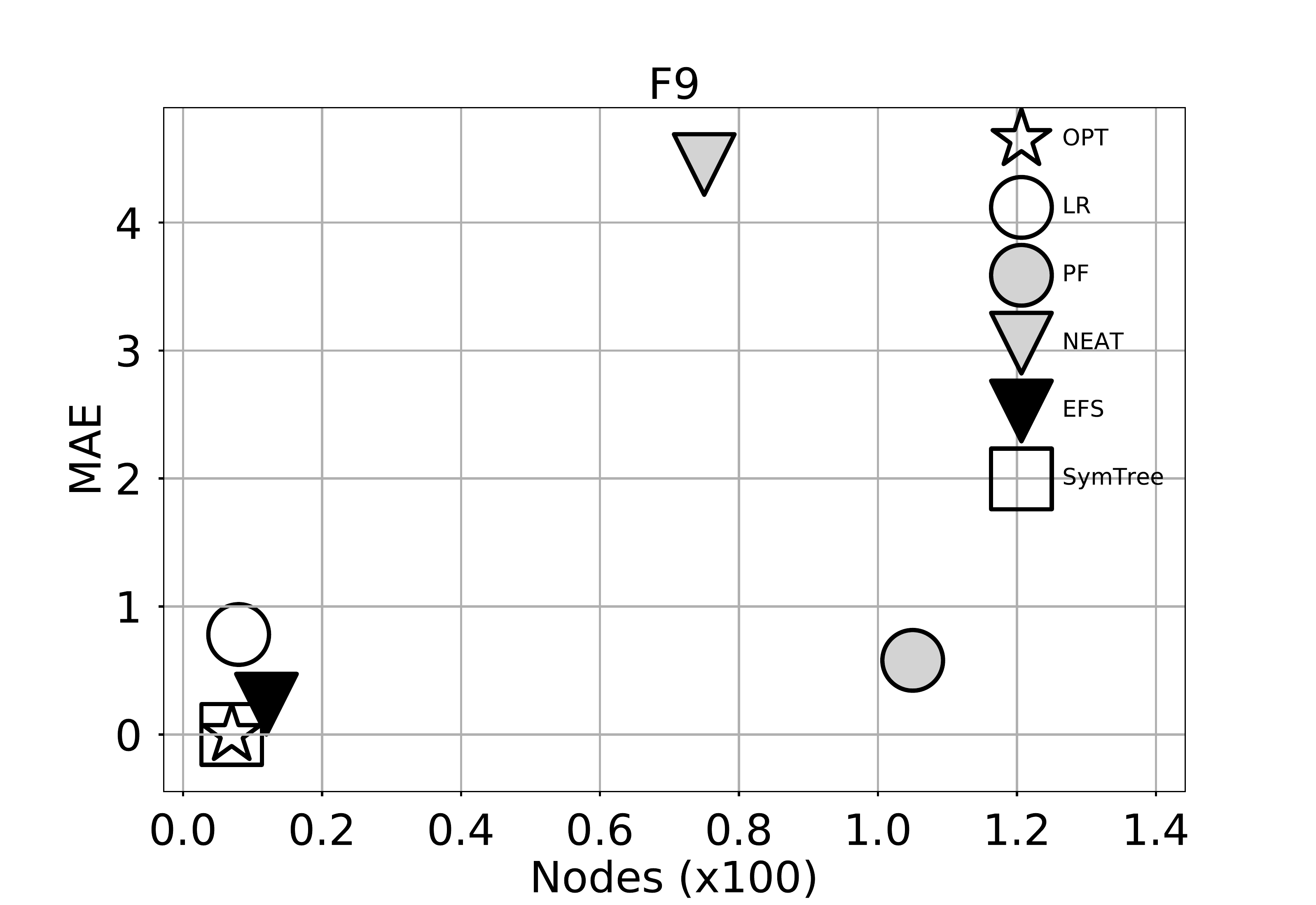}
  }
  \subfigure[]{
  \includegraphics[trim = 0mm 0mm 30mm 5mm, clip, width=0.31\textwidth]{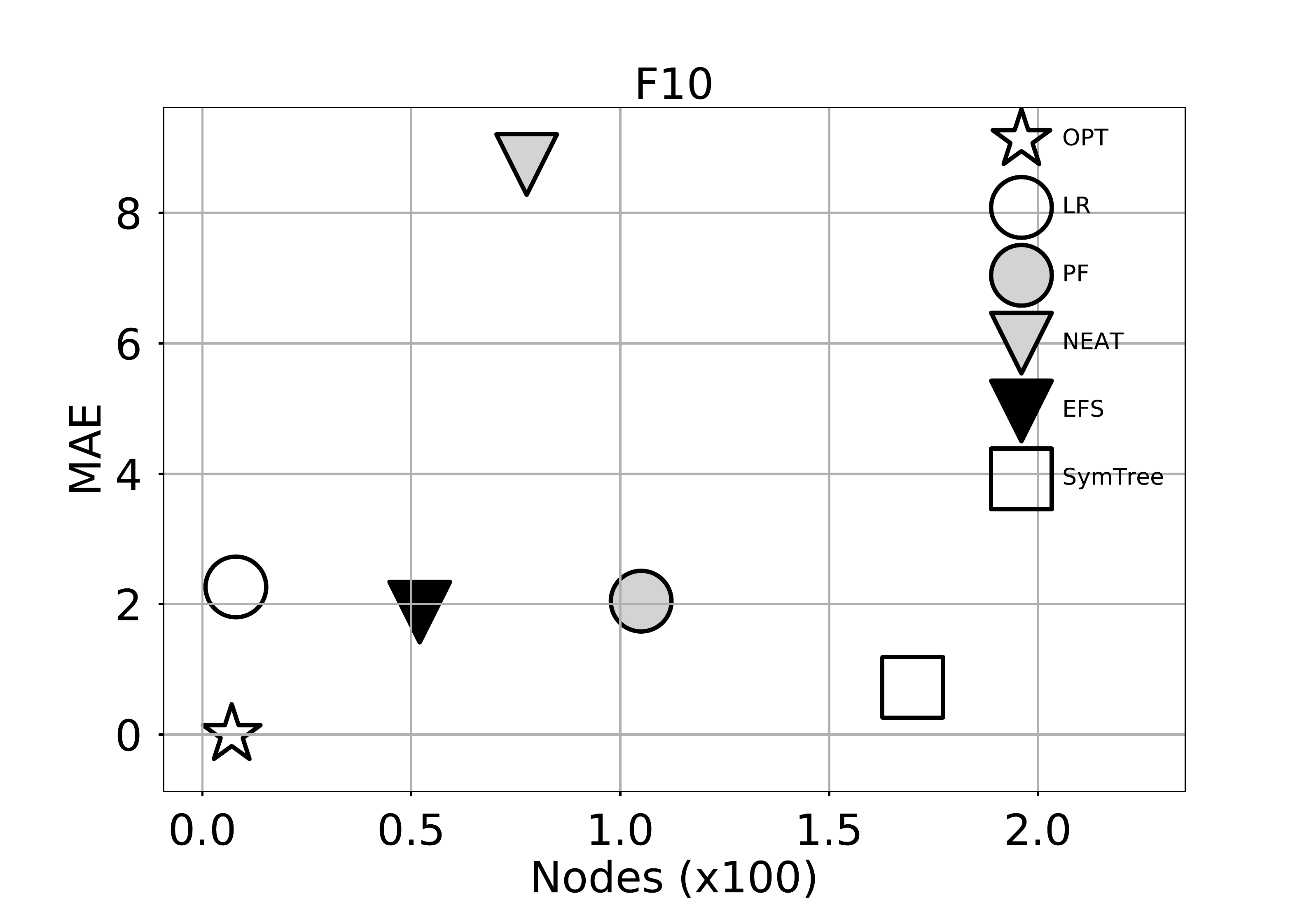}
  }
  \subfigure[]{
  \includegraphics[trim = 0mm 0mm 30mm 5mm, clip, width=0.31\textwidth]{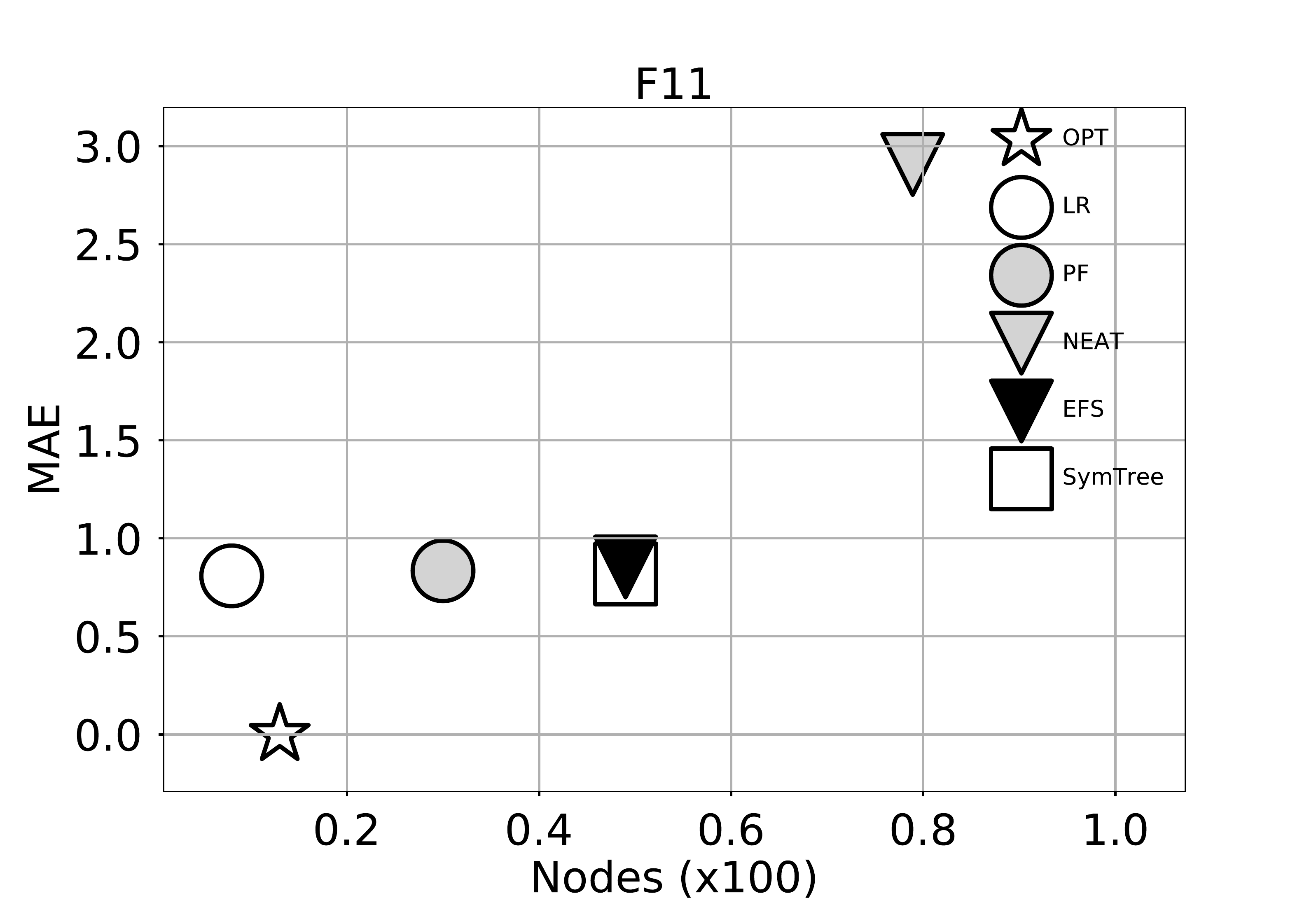}
  }
  \subfigure[]{
  \includegraphics[trim = 0mm 0mm 30mm 5mm, clip, width=0.31\textwidth]{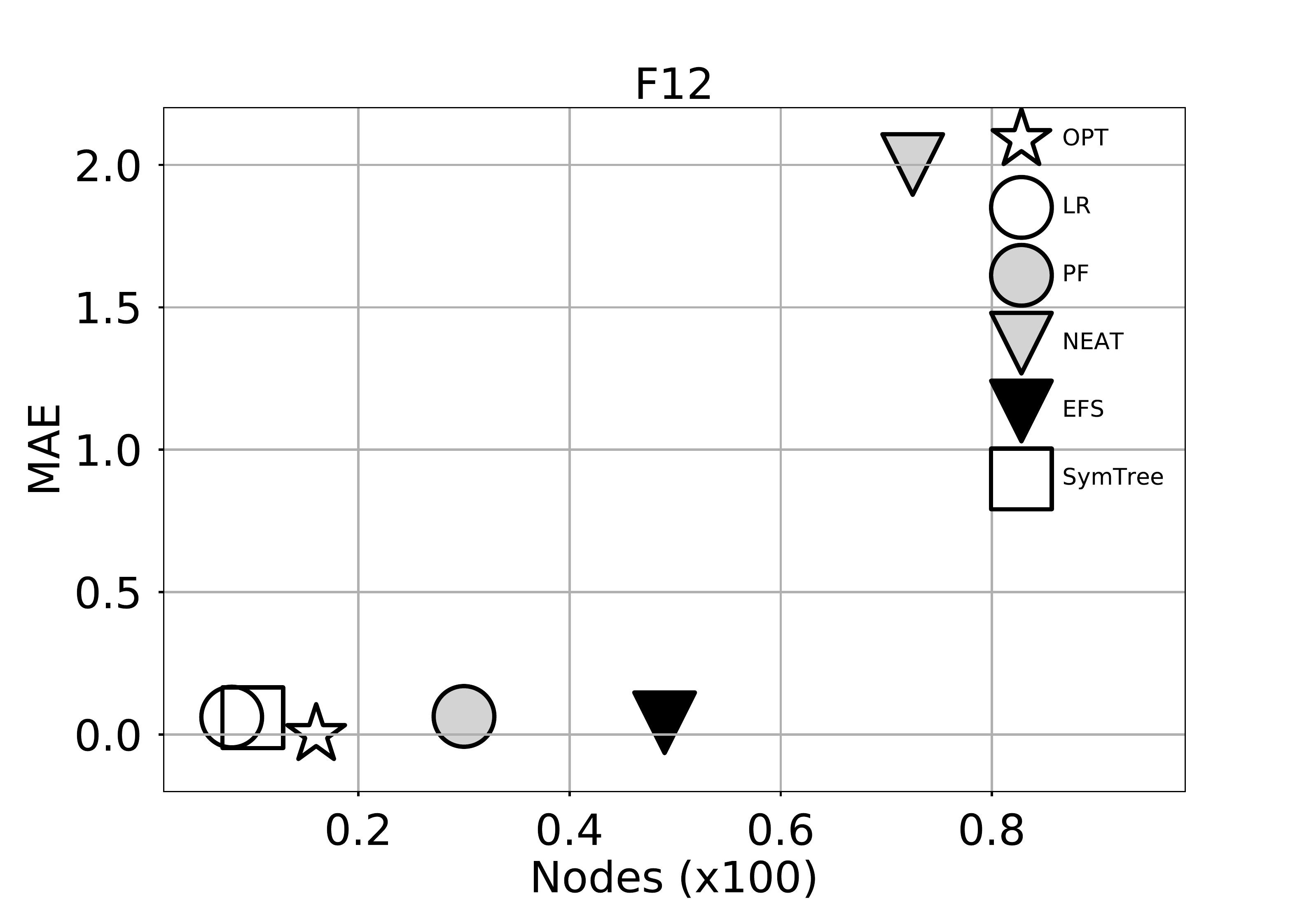}
  }

\caption{Compromise between accuracy and simplicity for functions F7 to F12.}
\label{fig:compromise2}
\end{figure}

\begin{figure}[ht!]
\centering
  \subfigure[]{
  \includegraphics[trim = 0mm 0mm 30mm 5mm, clip, width=0.31\textwidth]{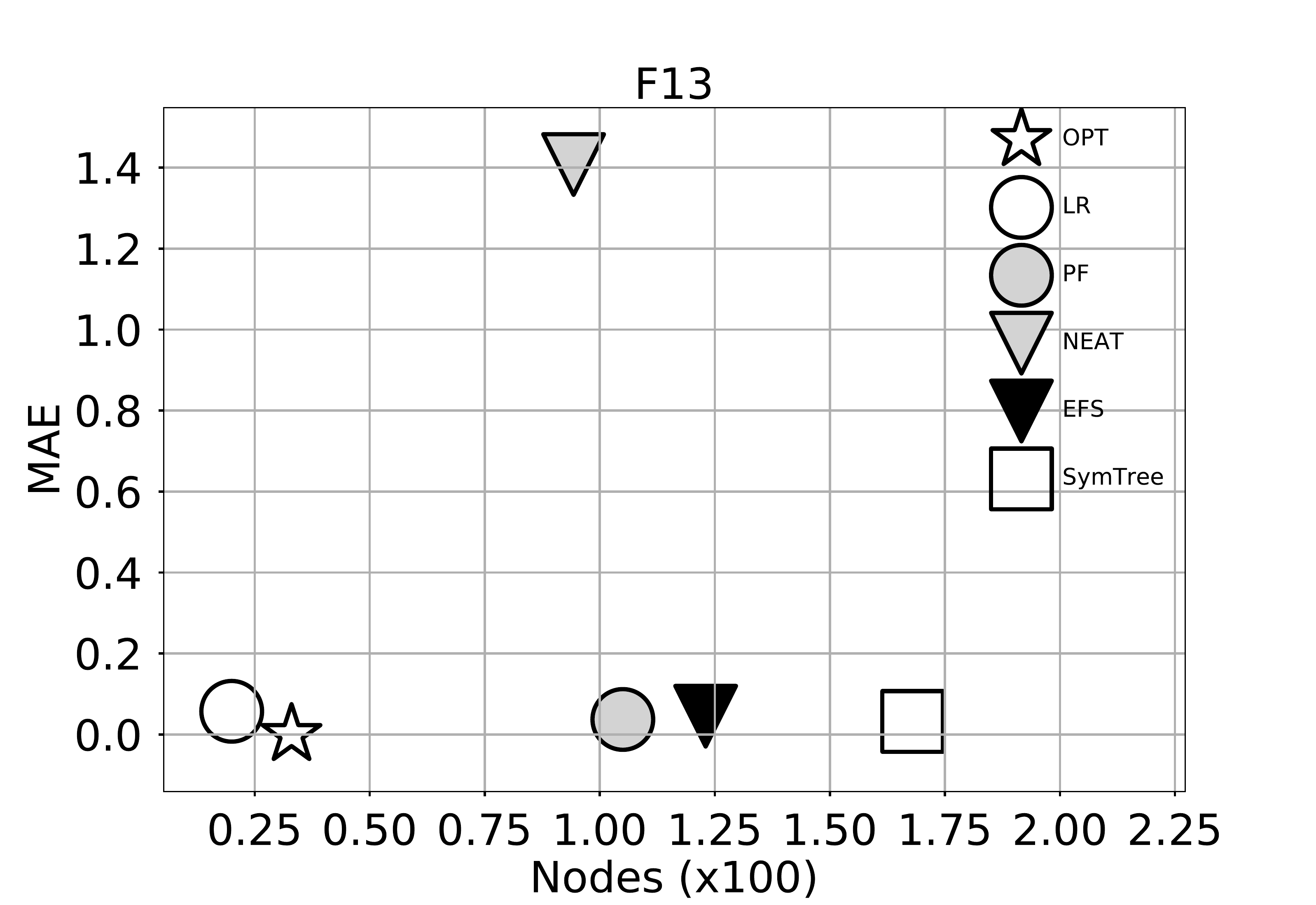}
  }
  \subfigure[]{
  \includegraphics[trim = 0mm 0mm 30mm 5mm, clip, width=0.31\textwidth]{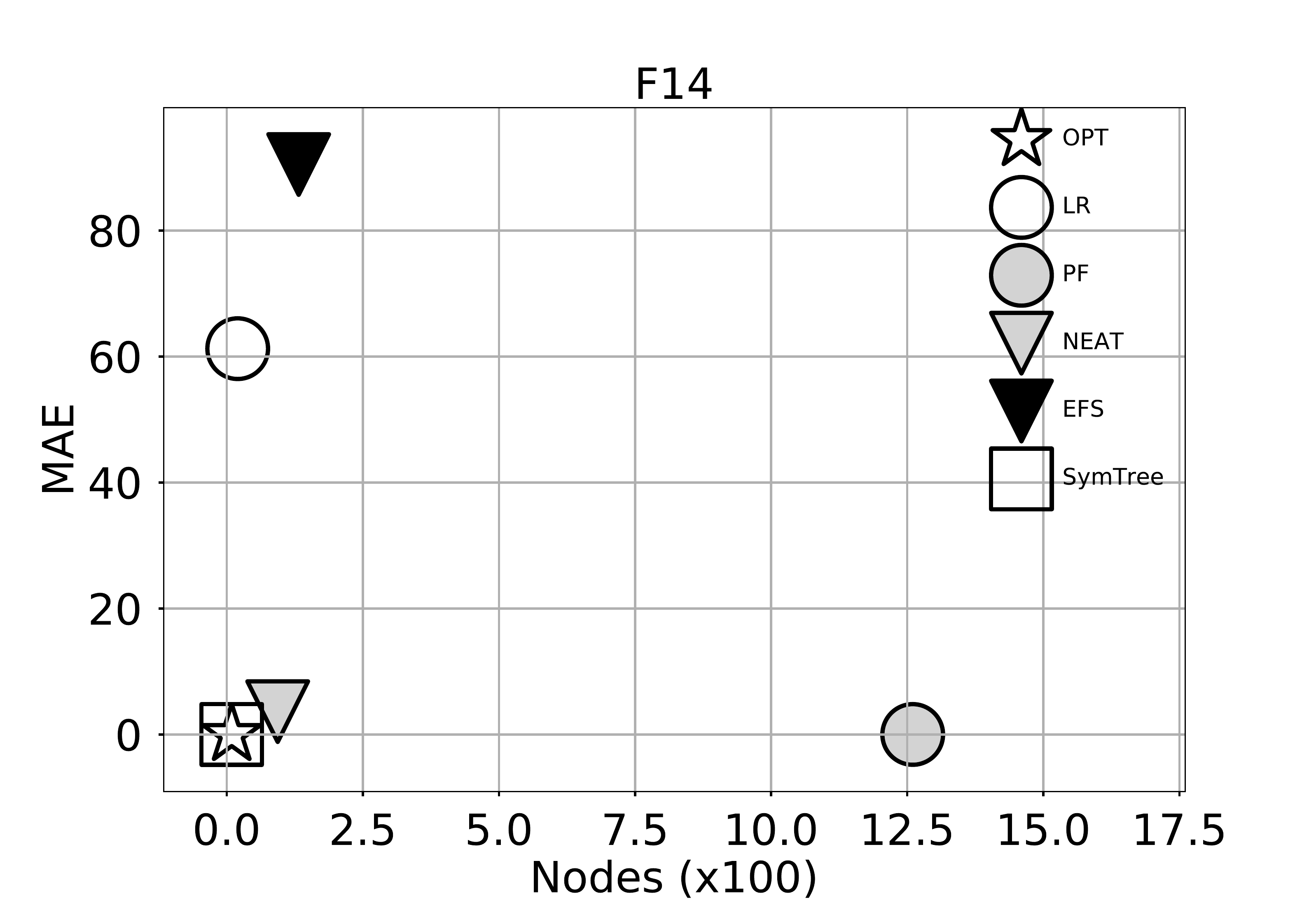}
  }
  \subfigure[]{
  \includegraphics[trim = 0mm 0mm 30mm 5mm, clip, width=0.31\textwidth]{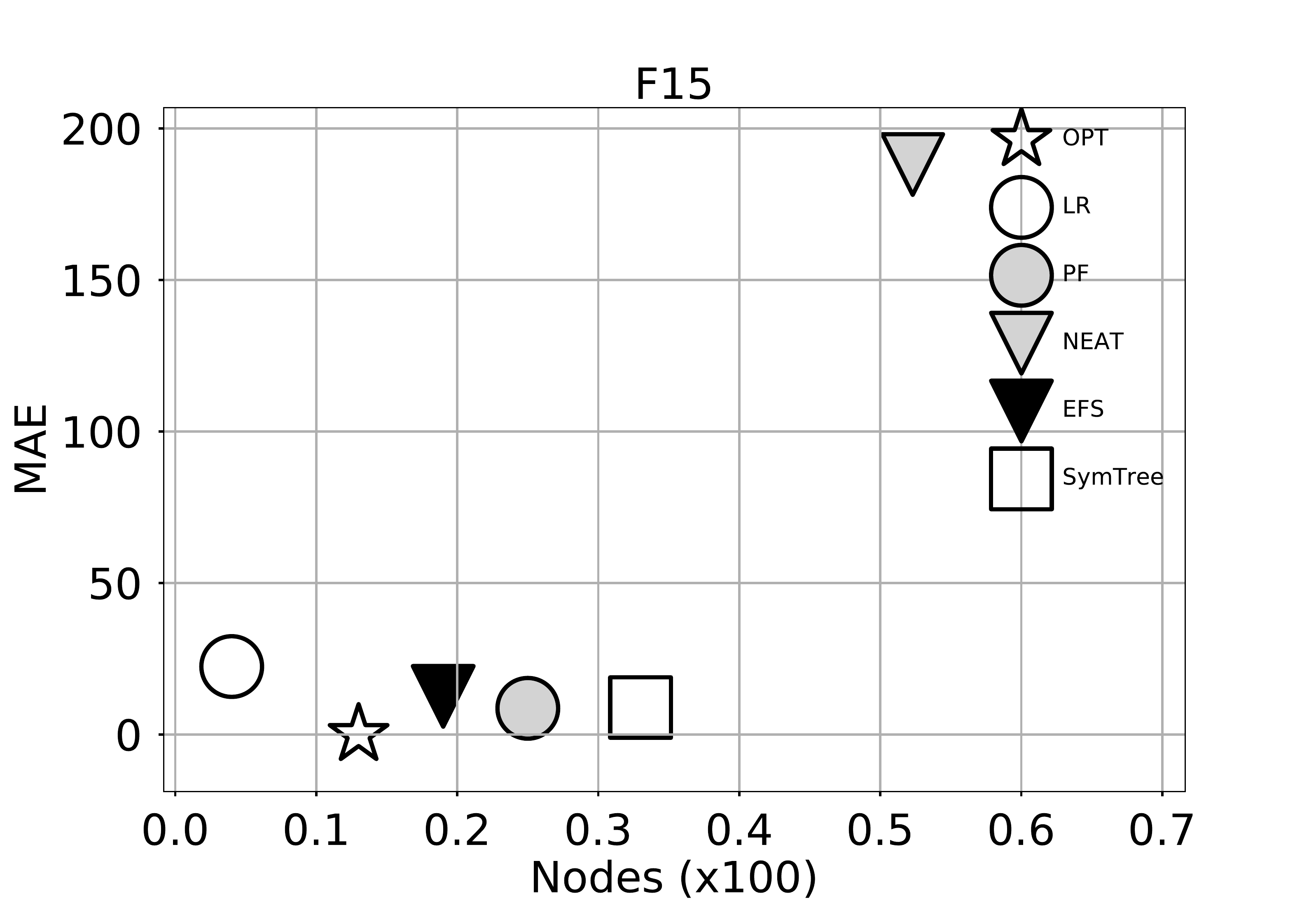}
  }
  \subfigure[]{
  \includegraphics[trim = 0mm 0mm 30mm 5mm, clip, width=0.31\textwidth]{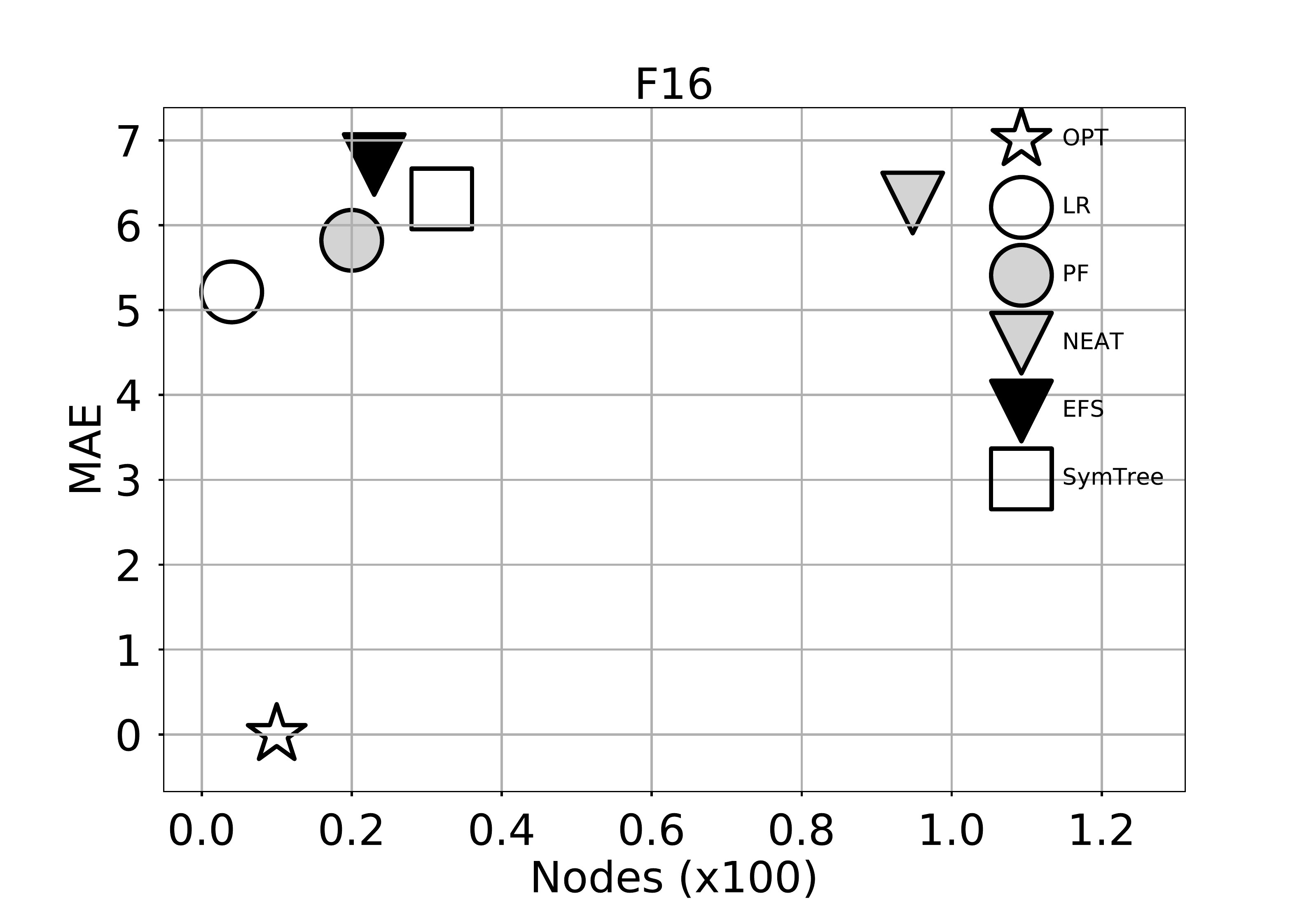}
  }
  \subfigure[]{
  \includegraphics[trim = 0mm 0mm 30mm 5mm, clip, width=0.31\textwidth]{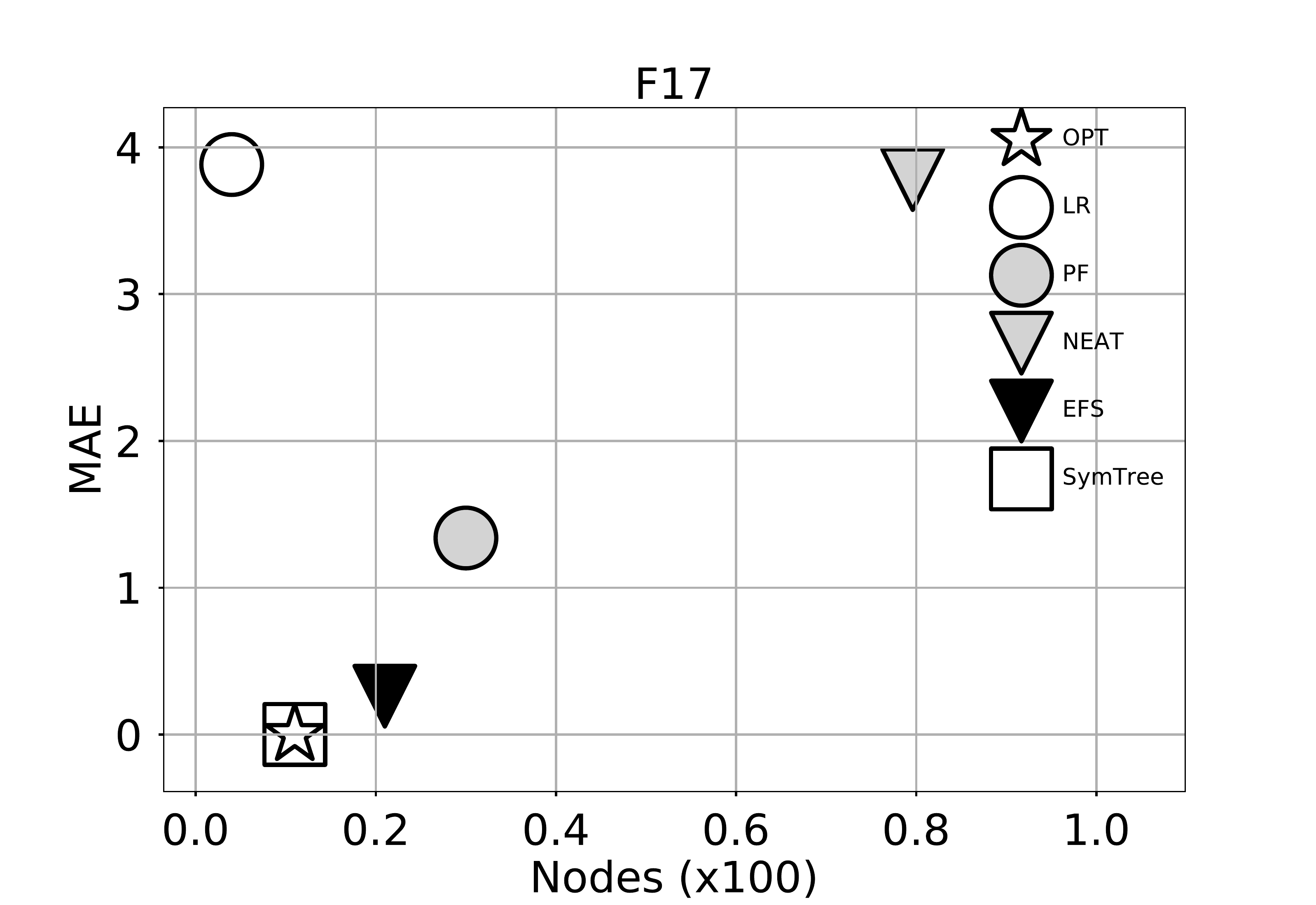}
  }

\caption{Compromise between accuracy and simplicity for functions F13 to F17.}
\label{fig:compromise3}
\end{figure}

In order to evaluate the compromise between accuracy and simplicity of the expressions generated by the algorithms we have plotted a scatter plot with the size of the expression tree obtained by each algorithm against the MAE on the testing data set. These plots are depicted in Figs.~\ref{fig:compromise1}~,~\ref{fig:compromise2}~and~\ref{fig:compromise3}. We have omitted the results obtained by GB in the following graphs since this approach generated a much larger expression (as a black box model) ranging from $1,200$ to $30,000$ nodes.

From these figures we can see some distinct situations. Whenever the target function could be represented by IT data structure, SymTree algorithm achieved the optimal compromise. In almost every situation, \emph{neat}-GP found a bigger expression with a higher error. The behavior of SymTree applied to those functions in which it had reached the optima does not clearly indicate whether it is biased towards generating simpler or more accurate expressions, but we can see that with the exception of four benchmark functions, it found a good balance between both objectives.

The generated expression by SymTree for each benchmark function can be found at~\url{http://professor.ufabc.edu.br/~folivetti/SymTree/}.


\subsection{Genetic Programming Based Symbolic Regression Using Deterministic Machine Learning}
\label{sec:deterministic}

As a final experiment, we compared the results obtained by FFX/GP algorithm~\cite{icke2013improving} on randomly generated polynomial functions. In this experiment, $30$ random functions for different combinations of dimension, order of the polynomial and number of terms were created. The dimension was varied in the set $\{1,2,3\}$, the order of the polynomial and the number of terms were varied within the set $\{1,2,3,4\}$. For every combination, it was generated $2500$ samples for training and $1500$ samples for testing all in the domain range $[0,1]$. Notice that every function generated on this test is achievable by SymTree. The results were reported by means of number of correct expressions found.

In this paper the authors also performed some tests with $10$ and $30$ dimensions, but the results were not reported on a Table, so we will refrain from testing them in this paper.

The results depicted in Table~\ref{tab:gpsr} show the number of correct polynomial expressions found by SymTree (number on the left) and FFX/GP (number on the right) for $1$, $2$, and $3$ dimensions respectively. The reported results for FFX/GP were the best values from the pure GP, FFX and the combination of FFX with GP~\cite{icke2013improving}, as reported by the authors. From these tables we can see that SymTree could achieve almost a perfect score with just a few exceptions, vastly outperforming FFX/GP in every combination.

\begin{table}[ht!]
\centering
\caption{Comparison of results obtained by SymTree (left) and FFX/GP (right) by number of correct answers.}
\begin{tabular}{cccccc}
\hline\hline
\textbf{Dim.} & \textbf{Order / Base} &     1 &     2 &     3 &     4 \\
\hline
& 1     &  30 / 30 &   -- &   -- &   -- \\
& 2     &  30 / 30 &  30 / 29 &   -- &   -- \\
1D & 3     &  30 / 30 &  29 / 27 &  30 / 19 &   -- \\
& 4     &  30 / 30 &  29 / 28 &  29 / 16 &  29 / 17 \\
\hline
& 1     &  30 / 30 &   -- &   -- &   -- \\
& 2     &  30 / 30  &  30 / 29 &   -- &   -- \\
2D & 3     &  30 / 30  &  30 / 22 &  30 / 15 &   -- \\
& 4     &  30 / 30  &  30 / 20 &  30 / 11 &  30 / 3 \\
\hline
& 1     &  30 / 30  &   -- &   -- &   -- \\
& 2     &  30 / 30  &  30 / 26 &   -- &   -- \\
3D & 3     &  30 / 30  &  30 / 28  &  30 / 14  &   -- \\
& 4     &  30 / 30  &  30 / 17  &  28 / 12  &  30 / 6  \\
\hline\hline
\end{tabular}
\label{tab:gpsr}
\end{table}


\section{Conclusion}
\label{sec:conclusion}

In this paper a new data structure for mathematical expressions, named Interaction-Transformation, was proposed with the goal of constraining the search space with only simple and interpretable expressions represented as linear combination of compositions of non-linear functions with polynomial functions. Also, in order to test this data structure, a heuristic approach was introduced to assess the Symbolic Regression problem, called SymTree. The heuristic can be classified as a greedy search tree method in which it starts with a linear approximation function and expands its nodes through the interaction and transformation of the parent expression by means of a greedy algorithm. 

This algorithm was tested in a set of benchmark functions commonly used in the Symbolic Regression literature and compared against the traditional linear regression algorithm, the linear regression with polynomial features, gradient boosting and some recent Genetic Programming variations from the literature. The results showed that SymTree can obtain the correct function form whenever the target function could be described by the representation. And in every function that it was not capable of finding the optima, it was capable of finding competitive solutions. Overall, the results were positive chiefly considering the greedy heuristic nature of the proposal.

Another interesting fact observed on the results is that SymTree tends to favor smaller expressions in contrast with black box algorithms which tends to favor accuracy over simplicity. It is shown that SymTree can find a good balance between accuracy and conciseness of expression most of the time. The evidences obtained through these experiments point to the validity of the hypothesis that the IT data structure helps to focus the search inside a region of smaller expressions and that, even though the search space is restricted, it is still capable of finding good approximations to the tested functions.

As a next step, we should investigate the use of the proposed representation in an evolutionary algorithm context, with the operators inspired by the operators introduced with the greedy heuristic. Since the greedy approach creates an estimate of the enumeration tree, it may have a tendency of exponential growth on higher dimensions which can be alleviated by evolutionary approaches.


\bibliography{SymbolicRegression}

\end{document}